\documentclass{article}
\usepackage[nonatbib, final]{nips_2017}

\usepackage[utf8]{inputenc} 
\usepackage[T1]{fontenc}    
\usepackage{hyperref}       
\usepackage{url}            
\usepackage{booktabs}       
\usepackage{amsfonts}       
\usepackage{nicefrac}       
\usepackage{microtype}      
\usepackage{float}
\usepackage{graphicx}
\usepackage{caption}
\usepackage{subcaption}
\usepackage[]{algorithm2e}
\usepackage{self}
\usepackage{times}
\usepackage{hyperref}
\usepackage{url}
\usepackage{natbib}
\usepackage{amsmath}
\usepackage{amsfonts}
\usepackage{amssymb}
\usepackage{graphicx}
\usepackage{multirow}
\usepackage{booktabs}
\usepackage{subcaption}
\title{A deep learning approach to real-time parking occupancy prediction in spatio-temporal networks incorporating multiple spatio-temporal data sources}

\author{
	Shuguan Yang\\
	Dept of Civil and Environmental Engineering\\
	Carnegie Mellon University\\
	Pittsburgh, PA 15213 \\
	\texttt{shuguany@cmu.edu} \\	
	\And
	Wei Ma\\
	Dept of Civil and Environmental Engineering\\
	Carnegie Mellon University\\
	Pittsburgh, PA 15213 \\
	\texttt{weima@cmu.edu} \\	
	\And
	Xidong Pi\\
	Dept of Civil and Environmental Engineering\\
	Carnegie Mellon University\\
	Pittsburgh, PA 15213 \\
	\texttt{xpi@andrew.cmu.edu} \\
	\And
	Sean Qian\\
	Dept of Civil and Environmental Engineering\\
	Heinz College\\
	Carnegie Mellon University\\
	Pittsburgh, PA 15213 \\
	\texttt{seanqian@cmu.edu}
}

\begin{document}

\maketitle
\begin{abstract}
	A deep learning model is applied for predicting block-level parking occupancy in real time. The model leverages Graph-Convolutional Neural Networks (GCNN) to extract the spatial relations of traffic flow in large-scale networks, and utilizes Recurrent Neural Networks (RNN) with Long-Short Term Memory (LSTM) to capture the temporal features. In addition, the model is capable of taking multiple heterogeneously structured traffic data sources as input, such as parking meter transactions, traffic speed, and weather conditions. The model performance is evaluated through a case study in Pittsburgh downtown area. The proposed model outperforms other baseline methods including multi-layer LSTM and Lasso with an average testing MAPE of 10.6\% when predicting block-level parking occupancies 30 minutes in advance. The case study also shows that, in generally, the prediction model works better for business areas than for recreational locations. We found that incorporating traffic speed and weather information can significantly improve the prediction performance. Weather data is particularly useful for improving predicting accuracy in recreational areas.
\end{abstract}
\section{Introduction}
Limited on-street parking availability and its associated traffic congestion have become one of the major issues of urban transportation systems. During peak hours, ``cruising for parking'' is a common phenomenon in areas with dense travel demand. Moreover, reliable sources for parking occupancy information, both current and forecast, and proper parking guidance are still lacking in most urban areas. As a result, the efficiency of travelers' searching for parking is significantly compromised due to the lack of information. It not only increases travelers' overall travel cost, but also adds additional congestion and emissions resulted from inefficient parking cruising. To tackle this issue, numerous parking management and guidance systems have been developed during the last few decades. A reliable source of predicted near-future parking occupancies is one of the key factors to their effectiveness. Only with a reliable prediction of on-street parking occupancies, proper parking location recommendations and navigation can be made in advance, followed by sharing the information to en-route drivers vis mobile devices or Advanced Driving Assistance Systems (ADAS). Moreover, through a reliable prediction of parking demand, the parking authorities can adjust the parking rate dynamically to improve the overall efficiency of the on-street parking resources\citep{sfpark,parkdc}, as well as to plan for their enforcement patrol accordingly.

In the literature, most studies ``estimate'' parking occupancies at the current moment \citep{boyles2015parking,yang2017turning,alajali2017street}, as opposed to ``predicting'' (or ``forecasting'') parking occupancies in the near future. There are generally two ways to predict short-term parking occupancy: (1) Model individual drivers' stochastic arrival and departure behaviors in a microscopic manner \citep{caicedo2012prediction,boyles2015parking,caliskan2007predicting}. The distributions of the arriving/departure process are commonly assumed as Poisson or Negative Exponential \citep{robert,arnott1995modeling}, and usually evaluated via simulations; (2) Model parking occupancy in data-driven approaches, which utilize statistical models trained by historical and real-time observations to predict aggregated parking occupancies in a mesoscopic manner, such as at the street or block levels \citep{tamrazian2015my}. In this study, we adopt the data-driven approach by incorporating multiple traffic-related sources, in terms of both real-time and historical data, including parking occupancy, traffic conditions, road characteristics, weather and network topology. It ultimately predicts (or forecasts) short-term parking occupancy via a deep neural network method.

In terms of the sources of parking data, most methods rely on data directly acquired from parking sensors of various kinds \citep{sfpark,hsu1989electronic,parkdc,parkMel,xiao2018likely}. Parking sensor data directly measure parking occupancies. However, there is lack of forecast of parking occupancies for the near future. The availability of sensing data depends on not only the deployment of sensors but also dedicated real-time communication systems, both of which are costly to build and maintain. To the contrary of the high cost of the sensing infrastructure is its relatively low reliability, due to issues related to sensor failure, error and power. Parking occupancies obtained directed from in-ground sensors has been reported not sufficiently accurate for real-time usage \citep{sfpark,parkMel}. In our method, we utilize parking meter transactions data instead of sensor data to first estimate the parking occupancy before the occupancy data is used for prediction (or forecast). This clearly offers cost advantage: since 95\% of the on-street paid parking are managed by meters, a prediction model based on transactions is more adaptive and cost efficient than relying on parking sensors. In our previous study, we have shown that the estimation of parking occupancy based on parking ticket transactions can be calibrated to a satisfying level given a small amount of ground truth data \citep{yang2017turning}.

Parking occupancies are related to many factors, such as travel demand, supply and weather. To best forecast parking occupancies in the near future, we would expect all those factors to play a role in understanding the possible occupancy change over time. Data sources that can be leveraged for parking occupancy prediction include traffic speed, traffic counts, incidents, physical supply (such as the number of lanes), and weather conditions, such as precipitation, wind speed and visibility. To illustrate the potential of utilizing those pieces of information in occupancy prediction, consider the following scenarios: during morning rush hours, if the traffic breakdown occurs later than usual, we infer that a portion of the daily commuters either leave home late, thus the peak of arrival parking is likely to be delayed as well (assuming everything else is the same as usual). For certain locations, such as museums or entertainment attractions, real-time traffic counts djacent roads, as well as the daily weather conditions, can partially reveal the parking demand in the near future. In case of a car accident or hazardous weather conditions on a weekday morning, travelers will be delayed due to slow traffic, thus also impacting the time-varying parking occupancy in the near future. In this paper, we attempt to collect all relevant data to comprehensively predict short-term parking occupancy for parking locations in networks.

In terms of the methods used for parking occupancy prediction, various statistical models are explored in the literature, including support vector regressions \citep{zheng2015parking}, clustering \citep{tiedemann2015concept,tamrazian2015my}, and time series models \citep{liu2010unoccupied}. There are also studies utilizing different types of simple neural networks\citep{rajabioun2015street,alajali2017street}. The most recently developed deep learning methods, such as Convolutional Neural Network (CNN) and Long-Short Term Memory (LSTM), have shown great promises in other prediction applications, but they have not been explored in parking prediction yet.
On the other hand, although numerous studies in the literature incorporate the spatio-temporal correlations of features of traffic conditions into their traffic prediction models \citep{kamarianakis2003forecasting,kamarianakis2012real,zou2014space}, parking occupancy prediction that also carries spatio-temporal features is overlooked yet. In fact, parking demand can exhibit certain spatio-temporal characteristics related to measures of traffic flow, incidents and weather. As a result, it is essential to model the spatial-temporal correlations among all the multi-sources data in order to most effectively predict parking usage.

To model the complicated spatio-temporal correlation among multi-source data and on-street parking occupancy, we used the combination of the graph based convolutional neural network (CNN) and recurrent neural network (RNN). Those methods are likely to encapsulate the non-linear relations to the full extent. The temporal relationships of parking occupancies along with other traffic-related data are simulated via Long-Short Term Memory (LSTM), while their spatial correlations are modeled through a graph-based CNN. The intuition can be interpreted as follows: During busy hours, the graph convolution operation can be interpreted as multiple drivers' cruising for parking behaviors among adjacent blocks. When a block's parking becomes full, parking occupancies on nearby blocks will be impacted more substantially than farther blocks. As a result, a multi-layered fully-connected decoder is used to parse the output from the previous network and generate the block-level occupancy prediction for each time period. 

This paper contributes to the literature in the following ways,
\begin{itemize}
	\item We propose a deep neural network based occupancy prediction model that utilizes multiple data sources, including parking meter transactions, traffic conditions, and weather conditions. While meter transactions data alone have been previously explored to estimate current parking occupancy \citep{sfpark,yang2017turning}, this study is the first to use parking meter transactions with other data sources to forecast (predict) short-term on-street parking occupancies.
	\item By combining Graph CNN and LSTM, the proposed structure is capable of modeling the spatial and temporal patterns of block-level parking occupancies with their relations to other data sources in the context of large-scale road networks. To our best knowledge, this is the first study that predicts parking occupancies on all blocks in a roadway network.
	\item We propose a general framework that is capable of incorporating any data input to the prediction model, including static data, graph-based, time series, as well as graph-based time series. The framework is flexible enough to cover essentially all types of data that is likely related to parking, providing the generality for the prediction model to be adapted in use cases.
	\item The performance of the method is evaluated with a case study in Pittsburgh downtown area. According to the result, the model outperforms other baseline methods including multi-layer LSTM, LASSO and historical mean, with a test error of 1.69 spaces in MAE when predicting block-level occupancies 30 minutes in advance. Moreover, the results prove the effectiveness of incorporating traffic speed and weather features into occupancy prediction. 
	\item We utilize multi-source data and explore their spatio-temporal correlation to predict parking occupancy. We examine the effectiveness of each data set in predicting real-time parking occupancy, and show that incorporating traffic speed and weather information can significantly improve the prediction performance. Weather data is particularly useful for improving predicting accuracy in recreational areas.
\end{itemize}
The rest of this paper is organized as follow: Section \ref{sec2} summarizes the literature review; Section \ref{sec3} introduces the proposed occupancy prediction model; Section \ref{sec4} presents the case study in Pittsburgh downtown area, including descriptions of the multiple data sets, results of the experiments and findings. Finally, Section \ref{cond} concludes the paper and discusses possible future work.

\section{Literature review}
\label{sec2}
Quite a few methods have been developed for estimating current parking occupancies and parking demand, \citet{elastic} discovered that parking occupancies estimated directly from aggregating transactions (namely every parker leaves exactly at the time ticket expiration) are generally lower than their actual observations, and such discrepancies are substantial in some cases. Similar conclusion was reached in a study comparing data from parking kiosks and actual counts for major paid parking blocks in Seattle \citep{sdot1, sdot2}, where the estimation error varies from 6\% to 55\%, and occupancies derived from transactions always being lower than the counted data. Other surveys \citep{heffron,robert,pierce2013getting} also discover that drivers' parking behavior is influenced by a set of external factors including time of day, day of week, parking duration, traffic congestion, weather and so on. As for predicting short-term parking occupancies, statistical and machine-learning models have been explored, including regression models based on prior assumptions on the distribution of parking occupancy \citep{pullola2007towards}, clustering \citep{tiedemann2015concept,tamrazian2015my}, multivariate spatio-temporal model \citep{rajabioun2015street}, chaos time series analysis \citep{liu2010unoccupied}, regression trees and support vector regression \citep{zheng2015parking}, autoregressive integrated moving average(ARIMA) \citep{burns1992econometric}, as well as continuous-time Markov Chain model \citep{klappenecker2014finding}.

During the last decade, models of deep neural networks are gaining popularity in the area of traffic modeling and prediction. Neural networks of various structures have been evaluated for the prediction of short-term traffic states. For example, \citet{ma2015large} proposed a two-layer restricted Boltzmann Machine(RBM) based RNN model to predict the congestion states for road links. \citet{lv2015traffic} adopted a five-layer stacked auto-encoder SAE to predict highway traffic flow. Stacked Restricted Boltzmann Machine mechanism is used by \citet{huang2014deep} for traffic flow prediction. Several studies have also examined short-term parking occupancy prediction. A three-layer wavelet neural network was used to predict short-term parking availability in off street garages \citep{ji2014short}. Multi-layer perceptron(MLPs) was used in predicting the parking occupancy of an individual parking location 1-30 minutes in advance \citep{vlahogianni2016real}. A representation learning method is adopted by \citet{ziat2016joint} to simultaneously predict traffic state and parking occupancies. A modified Neural Gas Network \citep{martinetz1993neural} is used for occupancy prediction on a single road segment in Berlin \citep{tiedemann2015concept}. A convolutional LSTM network is used by \citet{xingjian2015convolutional} to estimate precipitation. Later, \citet{li2017diffusion} applied a diffusion convolutional recurrent network structure to the prediction of short-term traffic, \citet{yu2017spatio} used gated convolution neural network to predict traffic in urban road networks. 
Most recently, In \citet{cui2018high}'s work, a high-order graph convolutional RNN is proposed for 5-min ahead traffic speed prediction in urban areas. This paper does not work with parking occupancy, but its methodology is closest to our work among all the relevant literature. While both this work and our paper share the idea of utilizing graph convolution operations to model spatial correlations, the essential differences reside in the way how a road network is incorporated in the prediction process. The road segment connectivities are converted to k-hop neighboring of the graph in their work, while our method takes a probabilistic way by using the shortest path driving time between parking locations when modeling the spatial correlations among users' cruising for parking behavior. In addition, rather than relying on a single-sourced data input, another key innovation of our method is the incorporation of multiple data sources (for both demand, supply and weather) with different spatial-temporal resolutions. Our work follows a general framework that is capable of incorporating any types of data input to the prediction, including static data, graph-based, time series, as well as graph-based time series.

In terms of data sources, very few prediction models have analyzed or incorporated traffic-related features other than parking occupancy itself. Pedestrians sensor data and daily average traffic counts are utilized previously without modeling the spatial correlations \citep{alajali2017street}. This study was for a particular location, since pedestrian counts are costly and challenging to obtain and thus hard to scale. In \citet{ziat2016joint}, traffic states are combined with parking occupancies as each data source is represented by time series of its own. Spatial correlations of traffic-related features among road links in networks are usually not explored, though they can help parking occupancy prediction tremendously. \citet{ziat2016joint} utilizes the network connectivity, time invariant information, to help parking prediction, and assumes that neighboring links share similar behaviors.

In this paper, we use a deep learning based prediction model that leverages graph convolutions to incorporate spatial-temporal features of multiple data sources acquired in networks, and characterizes of roadway networks related to parking cruising time. Previous research related to our methodology include structural-RNN \citep{jain2016structural} and Graph-based convolutional recurrent network \citep{seo2016structured}.

\section{Methodology}
\label{sec3}

\subsection{Motivation and model overview}
The objective is to extract as much information related to parking evolution as possible from external traffic-related data sources, such as traffic speed, traffic counts, public transit, incidents and weather. Under certain circumstances, those data, such as incident-induced congestion and snowstorms, can play a critical role in short-term forecast of traffic and parking demand. Multiple data sources are incorporate simultaneously in our model. Each data source are first embedded separately in our framework, and those embedded values are merged by a multi-layer decoder which outputs the predicted occupancies. As a result, our network based framework provides the flexibility that any individual data sources can be attached or detached from the neural network without compromising the overall structure of the prediction model. This improves the generalizability of the model since not all the aforementioned data sources are available for specific applications. Also, by evaluating the model performance under different combinations of input features, we can infer the effectiveness of each data source in occupancy prediction, and therefore identify the most effective models.

In the context of large scale road networks with multiple data sources, the dimension of data input to the model becomes too high for naive statistical methods. Thus, we propose a deep neural network by connecting graph CNN, LSTM and multi-layer feed-forward decoder. Such a structure can handle high dimensional space by modeling the non-linear correlations among spatial-temporal data and filtering out redundant information. Specifically, the spatial information is modeled through layers of convolutional neural network on graph (GCNN) \citep{niepert2016learning, henaff2015deep}. The GCNN uses the graph spectral theory \citep{chung1997spectral} to filter the signals on localized sub-graphs, then uses the filtered signals as the features for the neural networks. The temporal information could then be captured through a recurrent neural network (RNN), in our case, Long Short Term Memory (LSTM) \citep{hochreiter1997long}. Multiple data sources can be handled separately first by feature embedding and feature extractions for each data source. The extracted features are then combined as the input for a multi-layer decoder which yields the prediction of occupancies for each road link in the network.
Details of our model is discussed in the following subsections.



\subsection{Graph CNN}
The road network is modeled as a directed graph, with nodes being road link or parking locations, and edges transferring traffic flow or parking demand among nodes. To model the spatial correlations among road links or parking block in the road network, graph convolution operations are used to represent flow passage on the graph. Since nodes in a graph are not homogeneous to pixels in an Euclidean structure (such as an image), it is critical to define the local reception field in order to perform the convolution operations. Thus, we utilize spectral filters on signals \citep{niepert2016learning} to conduct convolutions on an directed graph.

Given a graph $G = (V, E, W)$, where $V$ is the set of nodes, $|V| = n$, $E$ is the set of edges and $W \in \mathbb{R}^{n \times n}$ is the weight matrix for all pairs of nodes. We define a signal $\boldsymbol{x} \in \mathbb{R}^n$ on the graph, where $\boldsymbol{x_i}$ is the signal for node $i$. Define the normalized Laplacian of the graph $L = I - D^{-\frac{1}{2}} W  D^{-\frac{1}{2}} $, where $D$ is the diagonal degree matrix with $D_{ii} = \sum_{j} W_{ij}$. The eigenvalues of $L$ are known as the frequencies of the graph, and the corresponding eigenvalues are known as graph Fourier modes. Thus by singular value decomposition, we have $L = U\lambda U^T$, where $\lambda = \texttt{diag}([\lambda_0, \cdots, \lambda_{n-1}])$, $U$ is the unitary eigenvector matrix.

From the perspective of road networks, the signal$x_i$ on a node is the observed value of the spatial data from one data source, such as parking occupancy or traffic speed. Each entry of the weight matrix $W \in \mathbb{R}^{n \times n}$ is the reciprocal of the historical average travel time between the center of the two road links on its shortest paths. Thus, the weight between two nodes in $W$ can be interpreted as their ``closeness'' measured by driving time. Naturally, road links close to each other tend to have higher correlation on parking occupancy or traffic congestion. When cruising for parking, travelers are more likely to explore nearby locations first than locations farther away.

We can then define a convolution operator on graph $G$ on the Fourier domain, the definition is presented in Equation~\ref{eq:gconv}.
\begin{eqnarray}
\label{eq:gconv}
\boldsymbol{y} = g_{\theta}(L)\boldsymbol{x} = U g_{\theta}(\lambda)U^T \boldsymbol{x}
\end{eqnarray}

We use $K$-order localized filters in the model so that the signals after convolution operations only incorporates information from their k-hop neighborhoods in the network. The localized filter shares the same ideas with tradition CNN on connectivity localization. The localized polynomial filter is presented in Equation~\ref{eq:lconv}. In which $\theta$ is the set of coefficients for the polynomial filtering, values of $\theta$ will be optimized during network training.
\begin{eqnarray}
\label{eq:lconv}
g_{\theta}(\Lambda) = \sum_{k=0}^{K-1} \theta_k \Lambda^k
\end{eqnarray}

To enhance the computational efficiency, we apply Chebyshev expansion to the polynomial filters \citep{kabal1986computation}. By Chebyshev expansion, terms in the polynomial filters can be expressed as:
\begin{eqnarray}
\label{eq:T_K}
T_k(\boldsymbol{x}) &=& 2xT_{k-1}(\boldsymbol{x}) - T_{k-2}(\boldsymbol{x})\\
\tilde{\Lambda} &=& 2\Lambda/\lambda_{max} - I
\end{eqnarray}
The polynomial filter can be reformulated as Eq~\ref{eq:fconv}.
\begin{eqnarray}
\label{eq:fconv}
g_{\theta}(\Lambda) = \sum_{k=0}^{K-1} \theta_k T_k(\tilde{\Lambda}) = \sum_{k=0}^{K-1} 2\theta_k \tilde{\Lambda} T_{k-1}(\tilde{\Lambda}) - \theta_kT_{k-2}(\tilde{\Lambda})
\end{eqnarray}
Comparing to the original approach shown in Equation~\ref{eq:lconv}, Chebyshev expansion eliminates the calculations of matrix multiplications, which significantly improve the computational efficiency in the case of  large scale road network (large $\tilde{\Lambda}$) and/or high-order polynomial filtering (large $K$).

\subsection{RNN and LSTM}
\begin{figure}[h]
	\centering
	\includegraphics[width=0.6\linewidth]{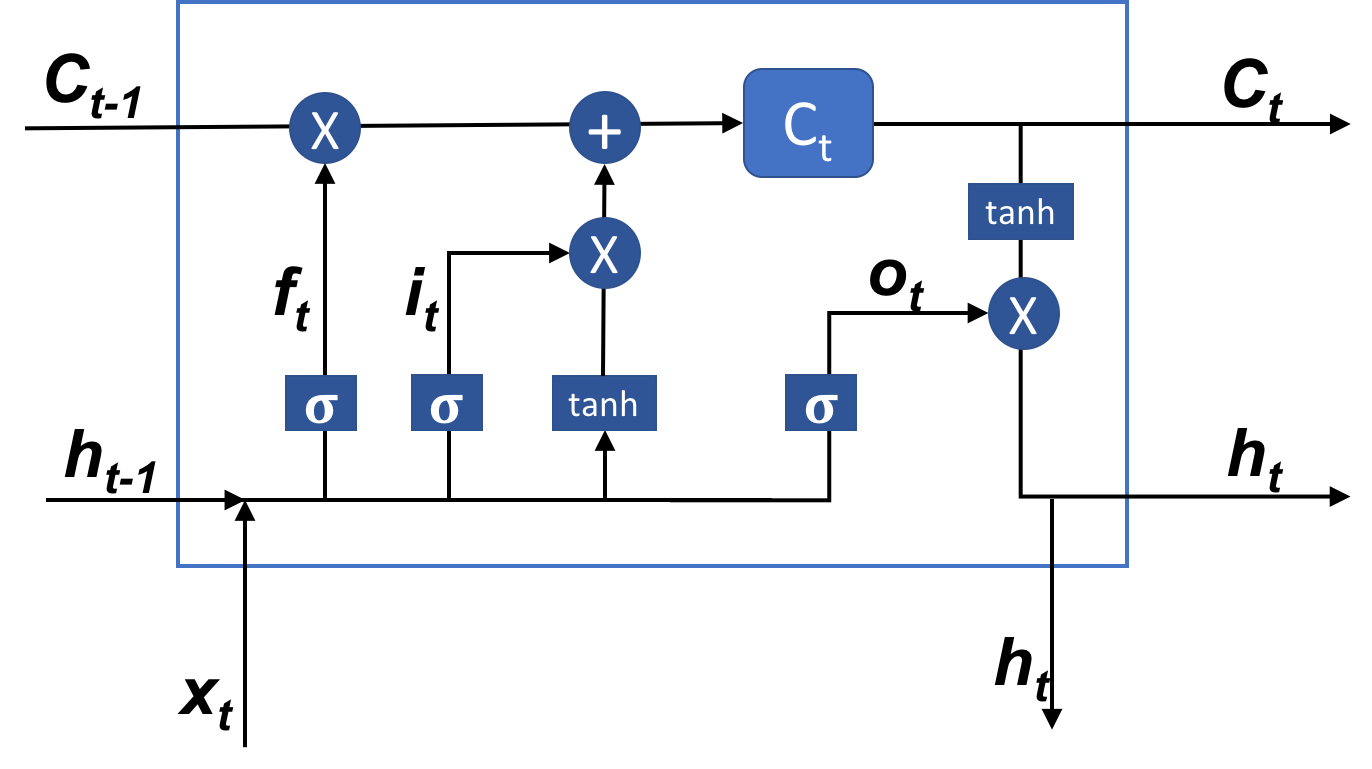}
	\caption{Scheme of an LSTM block \citep{graves2013speech}}
	\label{fig:lstm}
\end{figure}

Since various data sources for input are temporally correlated, such as speed, parking, public transit and weather, we utilize Recurrent Neural Network(RNN) which is a widely recognized deep learning approach to process temporal signals. We picked Long-Short-Term-Memory (LSTM) as the building block in RNN due to its success in numerous real-world applications. An LSTM cell is constructed with three gates, input gate $i_t$, forget gate $f_i$ and output gate $o_i$, and a memory cell $c_t$ is used to store the current state of the LSTM block. The formulation of LSTM is presented in Equation~\ref{eq:lstm}.

\begin{eqnarray}
\label{eq:lstm}
f_t &=& \sigma\left(W_f x_t + U_f h_{t-1} + b_f\right) \\ \nonumber
i_t &=& \sigma\left(W_i x_t + U_i h_{t-1} + b_i\right) \\ \nonumber
o_t &=& \sigma\left((W_o x_t + U_o h_{t-1} + b_o\right) \\ \nonumber
c_t &=& f_t \odot c_{t-1} + i_t \odot \mathrm{tanh} \left(W_c x_t + U_c h_{t-1} + b_c\right) \\ \nonumber
h_t &=& o_t \odot \mathrm{tanh}\left( c_t\right)
\end{eqnarray}
Where $x_t$ and $h_t$ are the input and output vector of the current time point, respectively, and $W, U, b$ being the weight and bias parameters of the block, and $\sigma$ is the sigmoid function. $\odot$ denotes the Hadamard product , or the entry-wise product of two matrices. The scheme of a single LSTM block is illustrated in Fig \ref{fig:lstm}.

The RNN framework can be used for sequence-to-sequence \citep{sutskever2014sequence} or sequence-to-one \citep{gers1999learning} learning. In our method, the sequence-to-one framework is used to predict the temporal features of the next time step. In order to make predictions of multiple time intervals ahead, we can simply forward the predicted values in RNN for any desired time intervals.

\subsection{Model Framework}
\begin{figure}[h]
	\centering
	\includegraphics[width=0.99\linewidth]{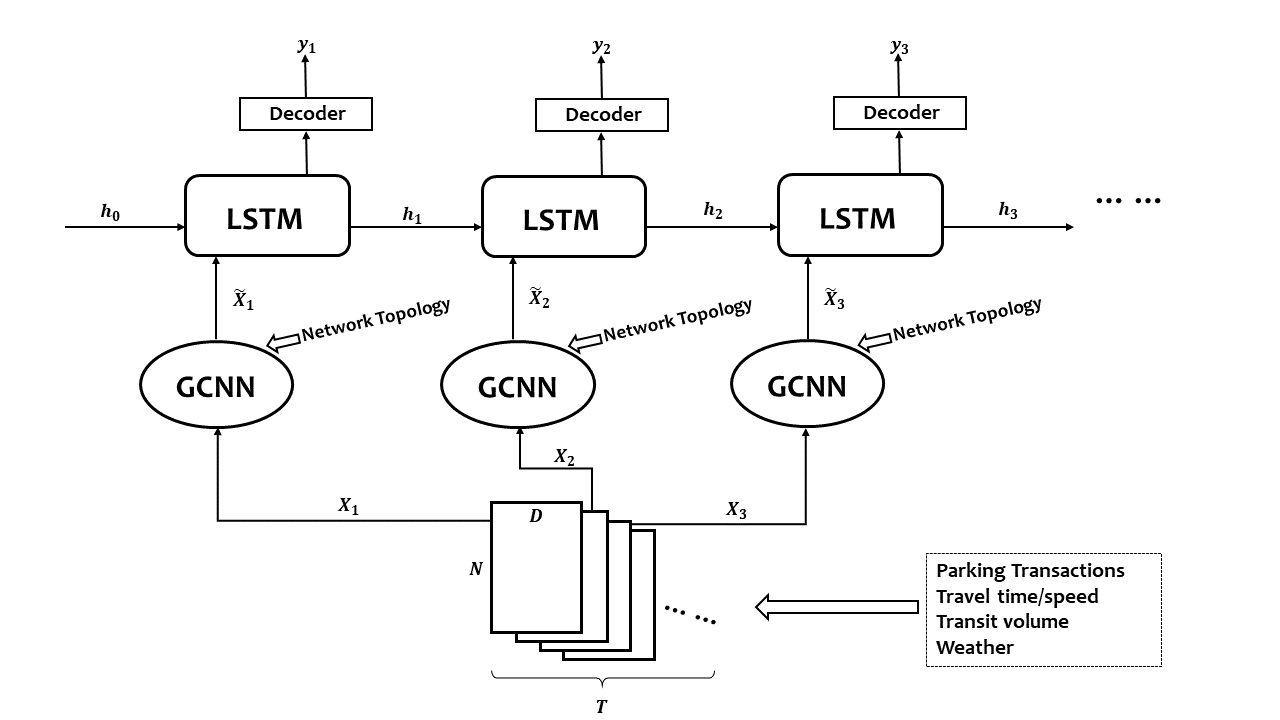}
	\caption{The network structure of the model}
	\label{fig:nn_structure}
\end{figure}

By combining the three network components, GCNN, LSTM and a decoder, sequentially, the proposed network structure of our model is illustrated in Figure \ref{fig:nn_structure}.

As stated before, data sources with different characteristics come in different formats. For example, speed and parking data has both spatial and temporal dimensions. Weather data for a single city only have  the temporal dimension. The number of lanes only varies spatially by road segments. Thus, our framework would provide the flexibility to take multiple data sources of different formats as input. Depending on the spatial and temporal characteristics, each data set first passes through the corresponding embedding layer(s) separately. For example, speed features pass through GCNN followed by LSTM, while weather features only pass through LSTM. Next, a combining layer aggregates all the features from each data set. A decoder of the multi-layer feed-forward network is then used to output the predicted occupancies at the level of city blocks.

In our framework, four data embedding schemes are developed coping for different type of data sources. These four data types are denoted as: $ND$, $NTD$, $NVD$, $NTVD$, where $N$ denotes the number of data, $D$ denotes the number of features, $T$ denoted the number of time stamps, $V$ denotes the number of nodes in the context of constructed graph.
\begin{itemize}
	\item $NTD$ dataset represents a traditional multi-variate time series dataset. The network-wise traffic state, temperature evolution can be modeled as $NTD$ datasets. We use sequence-to-one LSTM recurrent neural network to construct the embedding of this type of dataset. In order to archive the best performance, the number of LSTM layers as well as the dimension of each layer are tuned via grid search.
	\item $ND$ dataset does not contain time index nor spatial/graph information. Features such as local events and holidays can be interpretated as the $ND$ dataset. Influence of those features applies to the whole network and remains stable over time. We can use the traditional forward neural network to embed the $ND$ dataset. Non-linearity is also added to the embedding layer. In this paper, we adopt $\texttt{FC} \to \texttt{ReLU} \to \texttt{FC}$ as the embedding layer, where \texttt{FC} represents the fully connected inner product layer and \texttt{ReLU} is a Rectified Linear Unit as an activation function.
	\item $NVD$ dataset represents a graph-based dataset without temporal information, where the data are attached to different vertices in a graph. Daily average traffic volume, average link capacity flow rate can be modeled through a $NVD$ type of dataset. To extract features from graph-based dataset and consider the spatial relationship in the graph among features, we adopt Graph-CNN to compute the graph convolution to extract the higher order spatial information. Then we apply a \texttt{FC} layer to condense the spatial information from different vertices. The final embedded feature is of $NVO$, where $O$ is the embedded dimension for each vertex.
	\item $NVTD$ dataset represent a time-series data attached to different vertices in a graph. Real-time traffic speed data by road segment and real-time road closure data by road segment can be represented using a $NVTD$ dataset. The high cardinality of $NVTD$ dataset makes it hard to process and learn from. We proposed a parameter sharing technique to reduce the number of parameters in $NVTD$ embedding layer by a $NTD$ embedding layer and a $NVD$ embedding layer. We apply a graph-convolution filter to each time stamp, followed by an LSTM layer for the filtered data across the entire time horizon.
\end{itemize}

Apart from the flexibility of customizing the input space without changing the overall network structure, the proposed framework also enables network training in a distributed way, as embedding layers of individual data sources can be placed on separate GPUs during the training process.

\section{Case study: Pittsburgh Downtown}
\label{sec4}
To evaluate the performance of the proposed prediction model, we conducted a case study in Pittsburgh downtown area, which has 97 on-street parking meters scattered over the road network, as illustrated in the left figure of Fig\ref{Pit_map}. Four types of data are used in this case study, including parking meter transactions, traffic speed data, roadway networks, and weather conditions.

Data of all weekdays from 7AM to 6PM, during the period of 2014 Jan - 2015 Oct are acquired and used in this case study. All datasets are aggregated or interpolated into 10-min intervals. After removing invalid data points, there are in all 7,848 entries (namely the number of 10-min time intervals in the dataset) used in the case study. Among all the days in the dataset, 80\% of them are randomly selected as the training set, while the rest days are used as the testing set.  Based on the roadway network and on-street parking locations, the network is split into 39 street blocks, which become 39 nodes of a graph in our prediction model. The model is set to take data points of previous 4 hours as input, and output parking occupancy predictions of 30 minutes (3 10-min time intervals) in advance at the level of street blocks.

Data are explained in detail in the following subsections.

\subsection{Parking meter transactions}
\begin{figure}[h]
	\centering
	\includegraphics[width=0.48\linewidth]{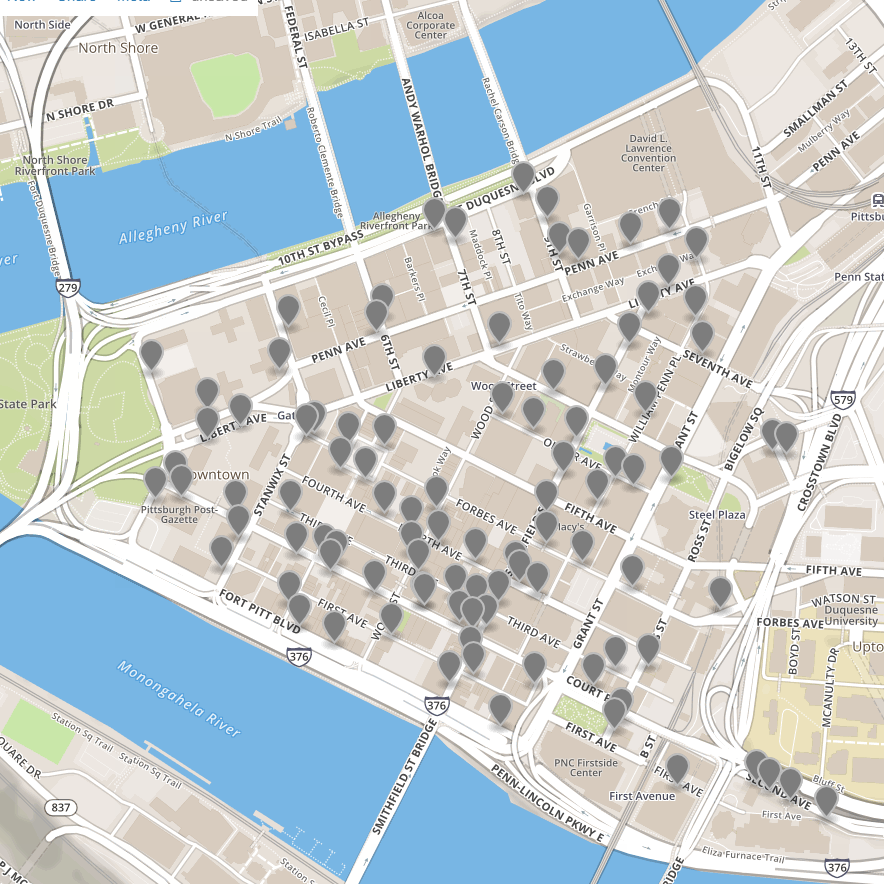}
	\includegraphics[width=0.48\linewidth]{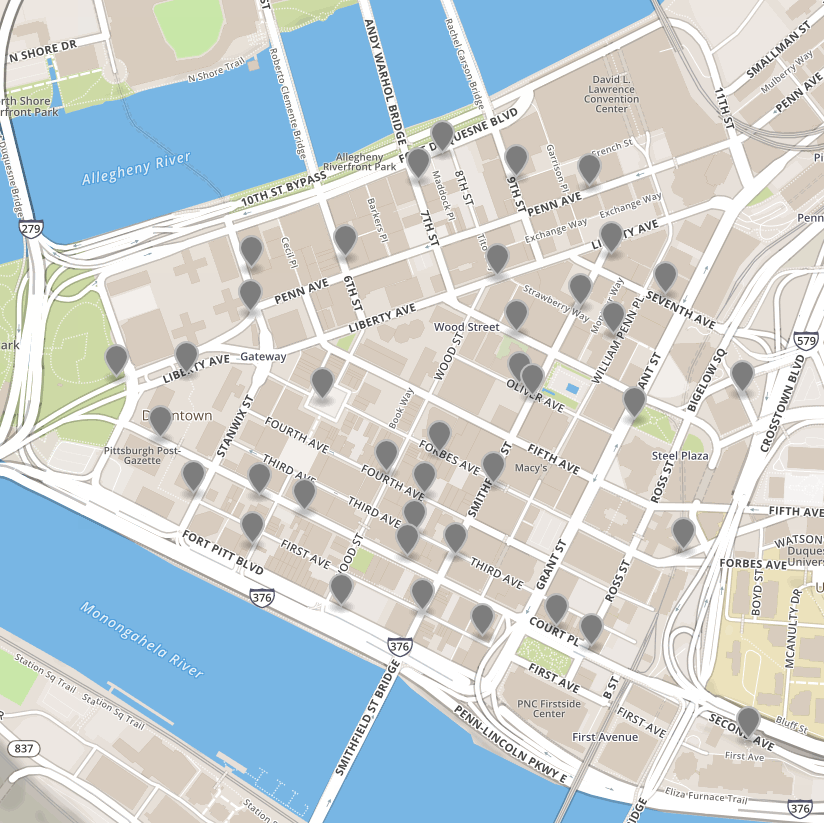}
	\caption{Left: Map of all parking meters around Pittsburgh downtown. Right: Map of the centroids of aggregated parking street blocks}
	\label{Pit_map}
\end{figure}

Historical parking transaction data are used as both model features and to compute occupancy. The transaction data is acquired from Pittsburgh Parking Authority. Each transaction entry comes with its start and ending time, meter ID and the parking rates. The parking transaction data exhibit recurrent patterns both by day of week and hour of day. There are in all 97 parking meters installed in the downtown area, and are aggregated into 39 street blocks based on their locations. Parkers are most likely to park in a block and purchase a ticket from any of the affiliated meters near this block. Locations of the 97 parking meters and centroids of the 39 street blocks are plotted in Fig \ref{Pit_map}.

Previously, we have shown that true parking occupancies can be approximated using meter transaction data with a few days of occupancy ground truth collected manually \citep{yang2017turning}. Since the main focus of this case study is evaluating the performance of the proposed model, we convert meter transactions to parking occupancies in a straight forward wa: we make the assumption that all vehicles start parking right after the payment is made and leave at the moment parking session expires, and there is no unpaid parking. Thus, the parking occupancy of a block at time $t$ is the aggregated number of transactions among all its parking meters that start before $t$ and expire after $t$. The goal is to predict this approximated parking occupancy 30 minutes in advance as parking transactions data are acquired in real time.

\subsection{Roadway network}
The roadway network defines the adjacency relationship as well as distances among different parking blocks. In this case study, we model the Pittsburgh downtown area with a directed graph of 39 nodes, where each node represent an on-street parking block. Two spatial data sources, parking transactions and speed data are pre-processed to fit this graph.

In our method, the network topology is processed as a distance matrix, where each entry is the reciprocal of the historical average travel time on the shortest path between the centroids of the two blocks. This information is obtained via the Google Map API\footnote{https://cloud.google.com/maps-platform/}. The distance matrix is then used as the node weight matrix $W \in \mathbb{R}^{n \times n}$ in the graph-CNN modules, which is used to filter signals in the spectrum domain.

\subsection{Traffic speed}
Traffic states, e.g. congestion level indicated by travel time and speed, may imply the current or near-future parking demand. Meanwhile, previous research has shown that a large portion of traffic congestion is caused by parking cruise \citep{2012trend}. Traffic speed is expected to help parking occupancy forecast.

The source of traffic speed in this case study is TMC (Traffic Message Channel) based data from INRIX. In addition to the real-time speed for each road segment, the dataset also provide the free flow speed as the reference speed. In this study, model features related to speed are processed as the free flow speed divided by the real-time speed, which can be interpreted as the level of congestion. This, comparing to the actual real-time speed, would be more effective in prediction, and thus the speed features all have the same scale across all road segments. TMCs are assigned to each parking block based on their overlap with parking street blocks. For those road segments without a TMC, a constant level of congestion as 1 is used. In this study, the temporal resolution of speed data is set as 10-minute.

The traffic speed has no ‘missing’ values as smoothing and other preprocessing are already conducted by INRIX. However, each speed entry comes with a confident score, ranging from 30 to 10, indicating the quality of the speed measurement. We removed entries with a confident score less than 30 from the dataset, namely all speed records used for training and testing are ‘observed’ value rather than ‘imputed’ values. 

\subsection{Weather}
Weather conditions may affect non-recurrent parking demand. In this case study, we obtain hourly weather reports of Pittsburgh downtown area via WeatherUnderground API\footnote{https://www.wunderground.com/}, and uses linear interpolations to smooth the weather data with a 10-min temporal resolution. In the input space, weather information for a time interval is an vector of floating or binary values of the following features: Temperature, Dew Point, Humidity, Wind speed, Wind gust, Visibility, Pressure, Wind chill temperature, Heat index, Precipitation intensity, Pavement condition, Fog, Rain, and Snow.

\subsection{Descriptive data analytics}
\label{data_ana}
Before applying the proposed model, we first conduct several descriptive experiments on the multi-source data, aiming to analyze the spatial and temporal characteristics of on-street parking in the network as well as its correlations with other data.

\begin{figure}[H]
	\centering
	\begin{subfigure}[b]{0.49\textwidth}
		\includegraphics[width=\textwidth]{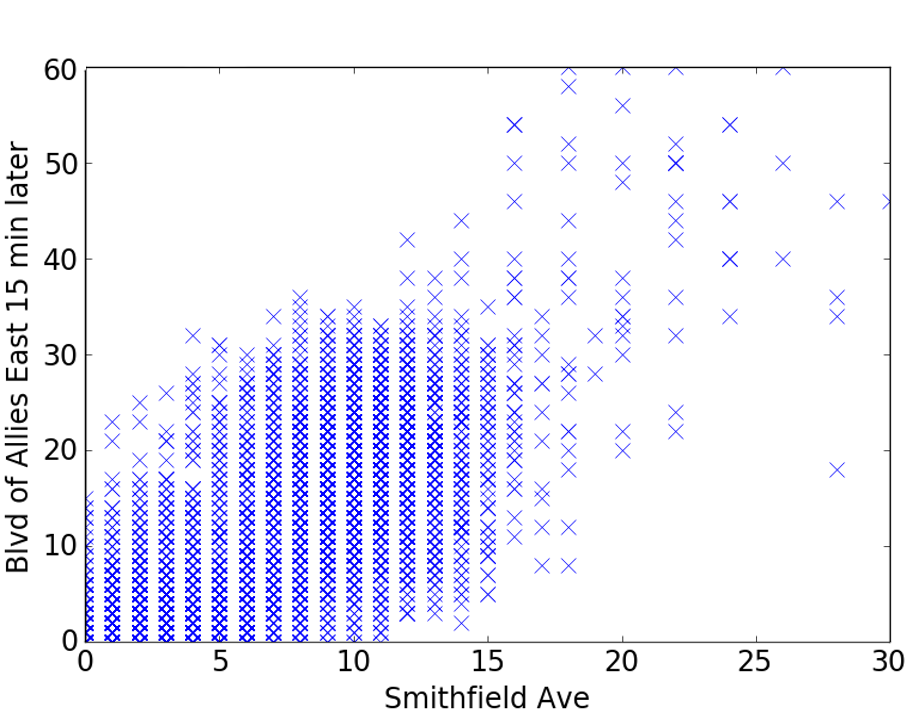}
		\caption{\footnotesize{Smith Ave versus Blvd of Allies E 15 mins later}}
	\end{subfigure}
	\begin{subfigure}[b]{0.49\textwidth}
		\includegraphics[width=\textwidth]{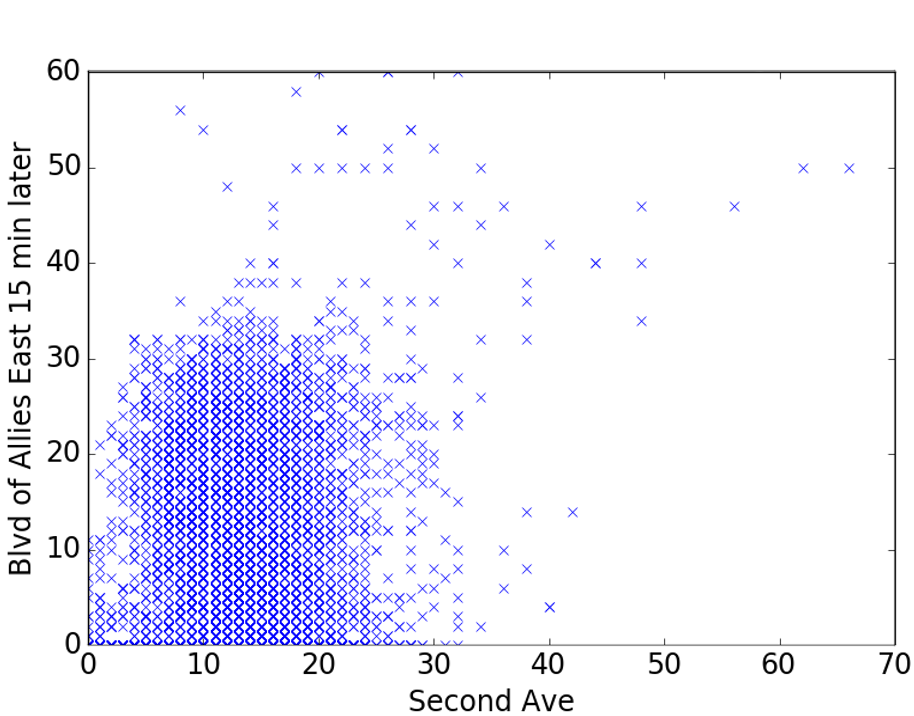}
		\caption{\footnotesize{Second Ave versus Blvd of Allies E 15 mins later}}
	\end{subfigure}
	\caption{An example of parking occupancy correlations among adjacent blocks in weekday morning (8 AM to 11 AM). The x and y axis are the parking occupancy measured in the number of parked vehicles of a parking block.}
	\label{fig:blks_com}
\end{figure}

To analyze the relationship between occupancies on adjacent parking blocks,  we visualize their correlations via 2D scatter plots. From the plots we discover that certain pairs of adjacent blocks owns strong linear correlations, while others do not. For example, correlations of parking occupancies on Boulevard of Allies East side, short as Blvd of Allies E, along with two adjacent blocks, Second Ave and Smithfield Ave, are plotted in Fig \ref{fig:blks_com}, for all weekdays from 8AM to 11AM. In addition, to test whether lagged temporal correlations exist, we applied a 15-min time lag for Blvd of Allies E. In other words, occupancies of the adjacent parking block are used to predict Blvd of Allies E's 15 minutes ahead occupancies. We find that occupancy on Smithfield Ave is strongly correlated with the Blvd of Allies E, while no significant correlation is found for Second Ave. Similar results are also found among other adjacent blocks. It looks like parkers' cruising may be limited to selected blocks only in an area, not all blocks. We infer that directly using network adjacencies only may not capture spatial correlations of parking demand or people's cruising for parking behavior. In addition, parking occupancy prediction models should be built and tuned for each specific location, not for all together.

\begin{figure}[H]
	\centering
	\begin{subfigure}[b]{0.49\textwidth}
		\includegraphics[width=\textwidth]{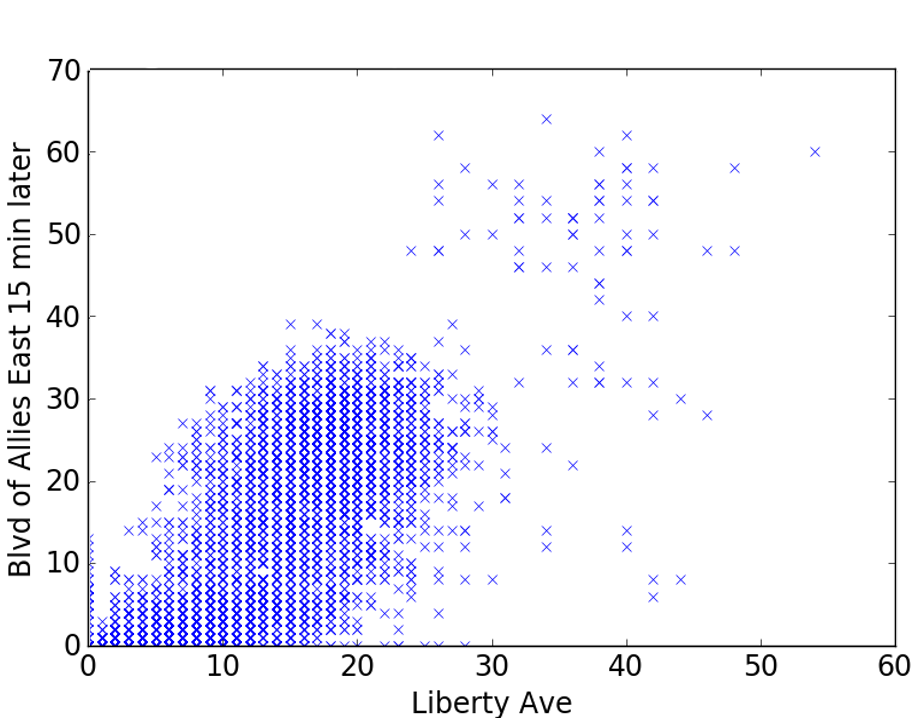}
		\caption{\footnotesize{Liberty Ave versus Blvd of Allies E 15 mins later}}
	\end{subfigure}
	\begin{subfigure}[b]{0.49\textwidth}
		\includegraphics[width=\textwidth]{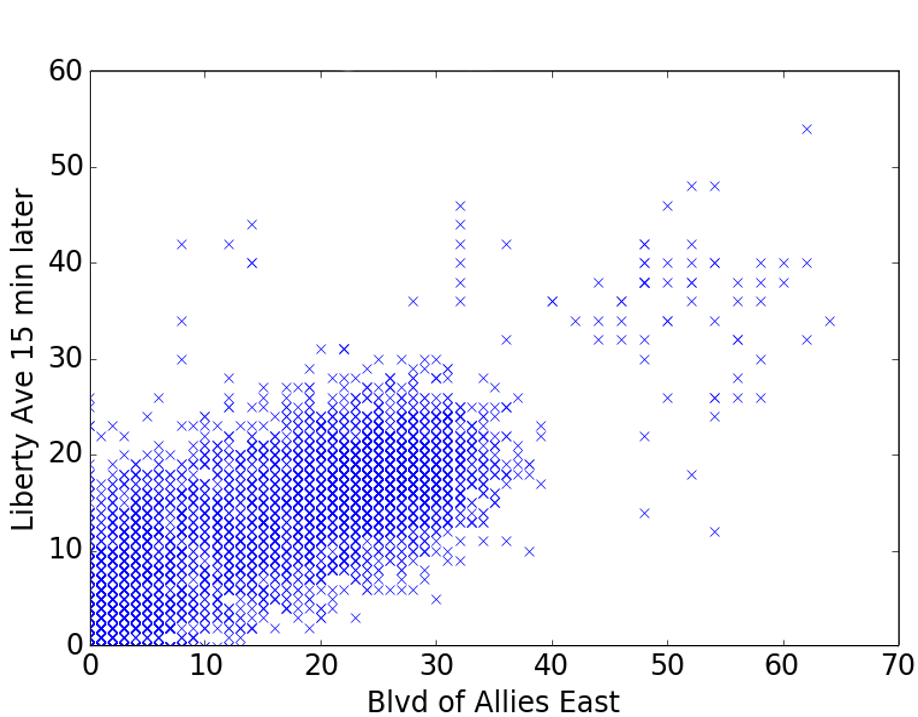}
		\caption{\footnotesize{Blvd of Allies E versus Liberty Ave 15 mins later}}
	\end{subfigure}
	\caption{An example of parking occupancy correlations among non-adjacent blocks in weekday morning (8 AM to 11 AM).The x and y axis are the parking occupancy measured in the number of parked vehicles of a parking block.}
	\label{fig:blk2blk}
\end{figure}

Next, we evaluate the correlations of occupancy among non-adjacent parking blocks in the road network. It turns out that strong correlations exist on many pairs of non-adjacent blocks. To demonstrate and conduct a direct comparison with the previous experiment for adjacent blocks, the Blvd of Allies E is used again as an example. In Fig \ref{fig:blk2blk}, strong linear correlations are shown between the Blvd of Allies E and Liberty Ave. Liberty Ave is distant from the Blvd of Allies E's in the network with approximately 15-min travel time. Parking occupancy on Liberty Ave owns a significant correlation with the Blvd of Allies E. Comparing to the results in Fig \ref{fig:blks_com}, the correlation is even stronger than two adjacent blocks of the Blvd of Allies E. On the other hand, by comparing the two plots in Fig \ref{fig:blk2blk}, we see that the correlations exist with a lag of 15 minutes along both directions. This can imply that parking cruising can start from one block to the other, and also the other way around. As measured by r2 values, the correlation between occupancy on Blvd of Allies E and Liberty Ave with a 15-min lag (right plot) is stronger than its opposite direction (left plot). This further illustrates the importance of adding spatial features of parking transactions in the network context when predicting parking occupancies of a single location.

\begin{figure}[H]
	\centering
	\begin{subfigure}[b]{0.49\textwidth}
		\includegraphics[width=\textwidth]{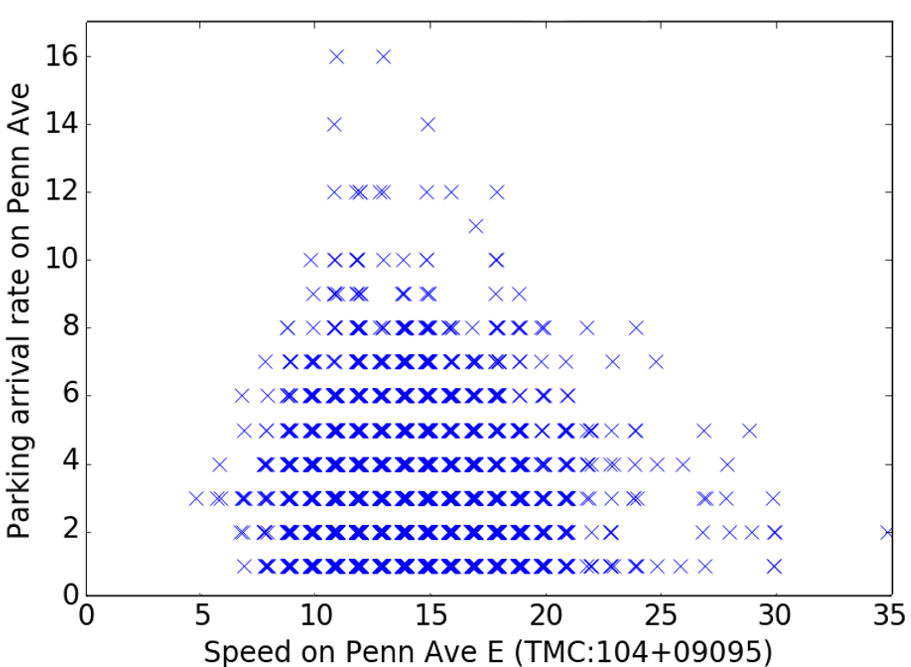}
		\caption{\footnotesize{Penn Ave East}}
	\end{subfigure}
	\begin{subfigure}[b]{0.49\textwidth}
		\includegraphics[width=\textwidth]{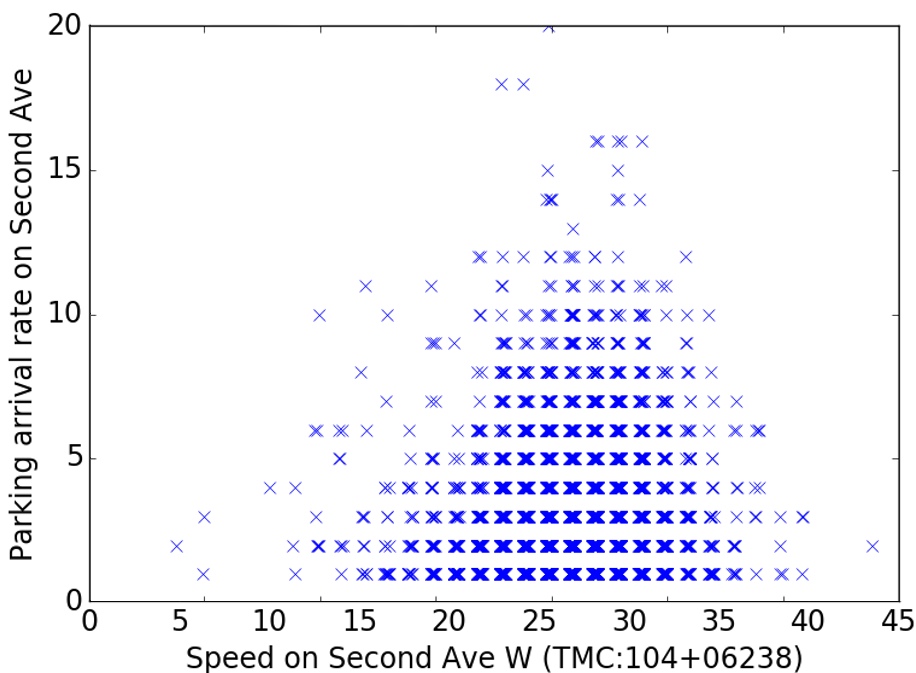}
		\caption{\footnotesize{ Second Ave}}
	\end{subfigure}
	\caption{Relationship between traffic speed and parking arrival rate on weekday non-rush hours(11 AM to 3 PM). In the plots, x axis is the speed measured in mph, y axis is the number of arrival vehicles in 15 min intervals.}
	\label{fig:spd2blk}
\end{figure}

We also analyze the relationship between parking occupancies and other traffic related data, including speed and weather. We use scatter plots to visualize the correlations between traffic speed and the parking arrival rate on each parking block. Two Examples are shown in Fig \ref{fig:spd2blk}, with the left plot for Penn Ave Eastbound and the right plot for Second Ave. The parking arrival rate is measured by the total number of vehicles arrived and parked on the block in each 15 min time interval. To remove the impacts of rush hours with a arrival rate peak, experiments are conducted on weekdays from 11 AM to 3 PM. It is interesting to see that there are clear, triangle-like plots. Parking arrival rates are usually high when speed is at its medium level, and low parking arrival rates can span a full range of traffic speed from free flow to congestion. Such a pattern could be explained as follows: In the cases that traffic demand is lower than usual, parking demand will also be lower, while travel speed will be around free-flow speed. In the cases of high traffic/parking demand, congestion occur and speed drops, and parkers need to spend more time on cruising and fewer of them can find parking spots given the low parking availabilities. When traffic speed is medium somewhere between free flow and congestion, a number of parkers are likely to easily find a spot, leading to substantial arrival rates.

Clearly, it is inappropriate to model the relationship between traffic speed and parking arrivals via models with low complexities, while a deep neural network may have the potential to learn this complicated relationship.

\begin{figure}[H]
	\centering
	\begin{subfigure}[b]{0.49\textwidth}
		\includegraphics[width=\textwidth]{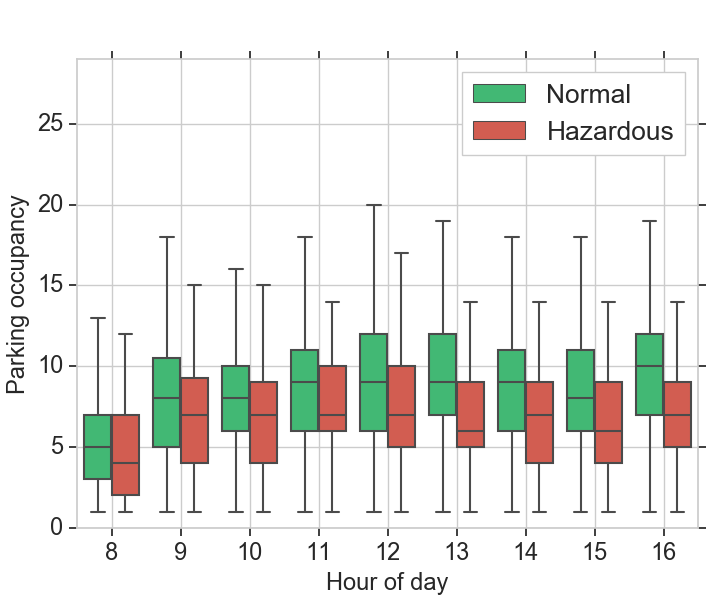}
		\caption{\footnotesize{Commonwealth Place}}
	\end{subfigure}
	\begin{subfigure}[b]{0.49\textwidth}
		\includegraphics[width=\textwidth]{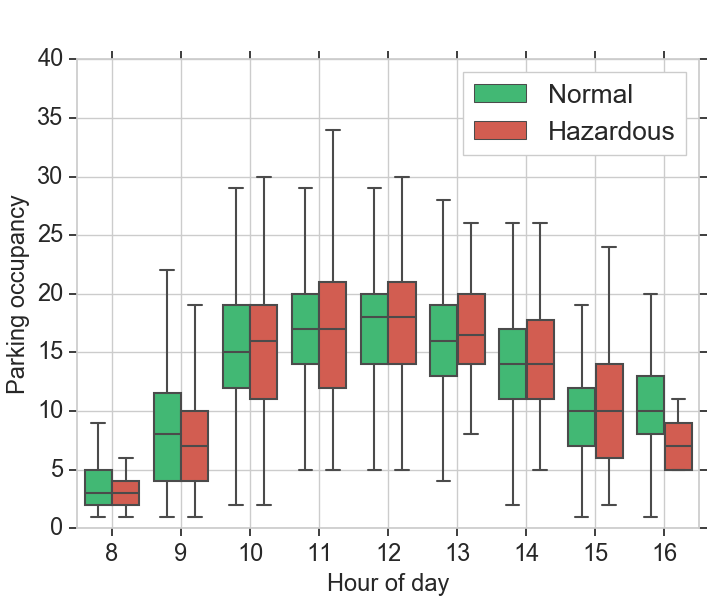}
		\caption{\footnotesize{ Liberty Ave}}
	\end{subfigure}
	\caption{Examples: Impact of hazardous weather on occupancies, weekdays all day (8AM to 4PM)}
	\label{fig:weather2blk}
\end{figure}

To evaluate the impacts of hazardous weather conditions on parking demand, such as raining, snowing and foggy, box plots are used to visualize the discrepancies between distributions of hourly parking occupancies under normal and hazardous weather conditions. Weather is considered hazardous if one or more of the following conditions are met: (1) Visibility is lower than 5 miles, (2) Precipitation intensity is higher than 0.15 inches/hour (3) snows. All other cases are considered normal weather conditions. The results of two representative parking blocks are shown in Fig \ref{fig:weather2blk}. As for Commonwealth Place, which is close to the Point State Park, significant drops in parking occupancies for all hours of day are observed under hazardous weather conditions; while such a pattern is not found on Liberty Ave, which is one of the business districts in downtown Pittsburgh. In fact, during mid-day and afternoon hours with hazardous weather conditions, average parking occupancies are even higher than normal on Liberty Ave, resulted from more people driving to work or make stops on days with hazardous weather. As a result, we can conclude that: (1) weather conditions impact parking occupancies, thus it is necessary to incorporate weather in the occupancy prediction model; and (2) the influence of weather conditions vary by location and time of day, attributing to the need to developing location-specific prediction models.

\subsection{Results and Discussions}
The proposed framework is implemented on PyTorch. All the experiments are conducted on a Linux workstation with two 1080Ti GPUs. All networks are trained by ADAM \citep{kingma2014adam} for 200 epochs or early stop if no improvement in 5 epochs. In each epoch, the entire training set is passed backwards through the neural network only once for parameter updates.

The performance of all networks are evaluated by the mean absolute error (MAE) as well as mean absolute percentage error (MAE) of predicting parking occupancies of all parking street blocks 30 minutes in advance. The definitions of MAE and MAPE are,
\begin{eqnarray}
\label{mse}
\textbf{MAE} = \frac{1}{V\times N}\sum_{i=1}^{V}\sum_{j=1}^{N}\left|y_{ij}-\hat{y}_{ij}\right|\\
\textbf{MAPE} = \frac{1}{V\times N}\sum_{i=1}^{V}\sum_{j=1}^{N}\left|\frac{y_{ij}-\hat{y}_{ij}}{q_{95}(\mathbf{y_{j}})}\right|
\end{eqnarray}
where $\hat{y}_{ij}$ is the predicted parking occupancies of block $j$ at time $i$ and $y_{ij}$ is the corresponding ground truth value. $\mathbf{y_i}$ is the vector of ground truth parking occupancies of all 39 blocks at time $i$, $V$ is the number of blocks predicted, namely $V = \left|\mathbf{y_i}\right|_0$, $N$ is the number of total observations. $q_{95}(\mathbf{y_{j}})$ is the 95th percentile of the ground-truth parking occupancies on block $j$, which is used as the approximated parking capacity of the block. For on-street parking locations in Pittsburgh downtown, parking spots are not marked, in which case there are no fixed ‘capacities’ for those blocks. The max number of spots on each block depends on the length of each parked car and the gaps between cars. As a work around, we calculate the distribution of the total number of cars parked on each block among all time slots, and use the 95th percentile as a ``reference capacity''. Another reason for not using the max occupancy as the capacity is that, due to unpaid and under-paid parking, the maximum occupancy directly derived from transactions data does not always reflect the true capacity. Overall, we use the 95th percentile occupancy to represent the parking capacity for each block.

To find the best model in this case study, we conduct grid searches on the hyper-parameter space of the network, hyper-parameters that are tuned include: 1) data preprocessing method: standardization, normalization, winsorization and MinMax Scaling on $[-1,1]$; 2) configurations of the network for each dataset: the number of layers, the number of neurons per layer, dropout rate; 3) activation functions: ReLu and sigmoid, which are defined in Eq \ref{reLU} and Eq \ref{Sigmoid} respectively; and 4) training strategy: base learning rate and momentum rate.
\begin{equation}
\text{ReLU:} \ f(x)=x^{+}=\max(0,x)
\label{reLU}
\end{equation}

\begin{equation}
\text{Sigmoid:}  \ f(x)={\frac {e^{x}}{e^{x}+1}}
\label{Sigmoid}
\end{equation}

After tuning all the hyper-parameters via grid search, the specifications of the final model are shown below:
\begin{itemize}
	\item {\em Data preprocess pipeline}: Normalization $\to$ MinMax Scaling.
	\item {\em Occupancy embedding}: $\texttt{GCNN} \to \texttt{FC1} \to \texttt{FC2} \to \texttt{LSTM}$. Dimension of \texttt{GCNN} output layer: $8\times V$. Dimension of \texttt{FC1} and \texttt{FC2}: $16\times V$ with dropout rate: $0.25$. Dimension of \texttt{LSTM} layer: $256$.
	\item {\em Speed embedding}: $\texttt{GCNN} \to \texttt{FC1} \to \texttt{FC2} \to \texttt{LSTM}$. Dimension of \texttt{GCNN} output layer: $4\times V$. Dimension of \texttt{FC1} and \texttt{FC2}: $8\times V$ with dropout rate 0.0. Dimension of \texttt{LSTM} layer: $256$.
	\item {\em FNN decoder}: $\texttt{FC1} \to \texttt{ReLU} \to \texttt{FC}$. Dimension of \texttt{FC1}: $1024$. Dimension of \texttt{FC2}: $V$.
	\item {\em Optimizer}: ADAM learning rate $0.001$, momentum rate: 0.9, weight decay (L2 regularization): 1e-4.
\end{itemize}
Where dimension is short for the dimension of the layer, $V$ is the total number of nodes in the graph, i.e., $V=39$. 
To alleviate overfitting during model training, an L2 regularization term is added to the loss function, which penalizes the model based on the amplitude of parameters and encourages less complex networks\citep{zhang2016understanding}, the weight of the regularization is set as $10^{-4}$. Another popular technic to restrict overfitting in neural networks is early stopping\citep{prechelt1998early}, which have been proven to improve the robustness of deep learning models trained by gradient descent. An early stopping strategy with $patience = 5$ is utilized in this study, the training is terminated once no improvement in testing is observed within the last $5$ epochs.

\begin{figure}[H]
	\centering
	\includegraphics[width=0.8\textwidth]{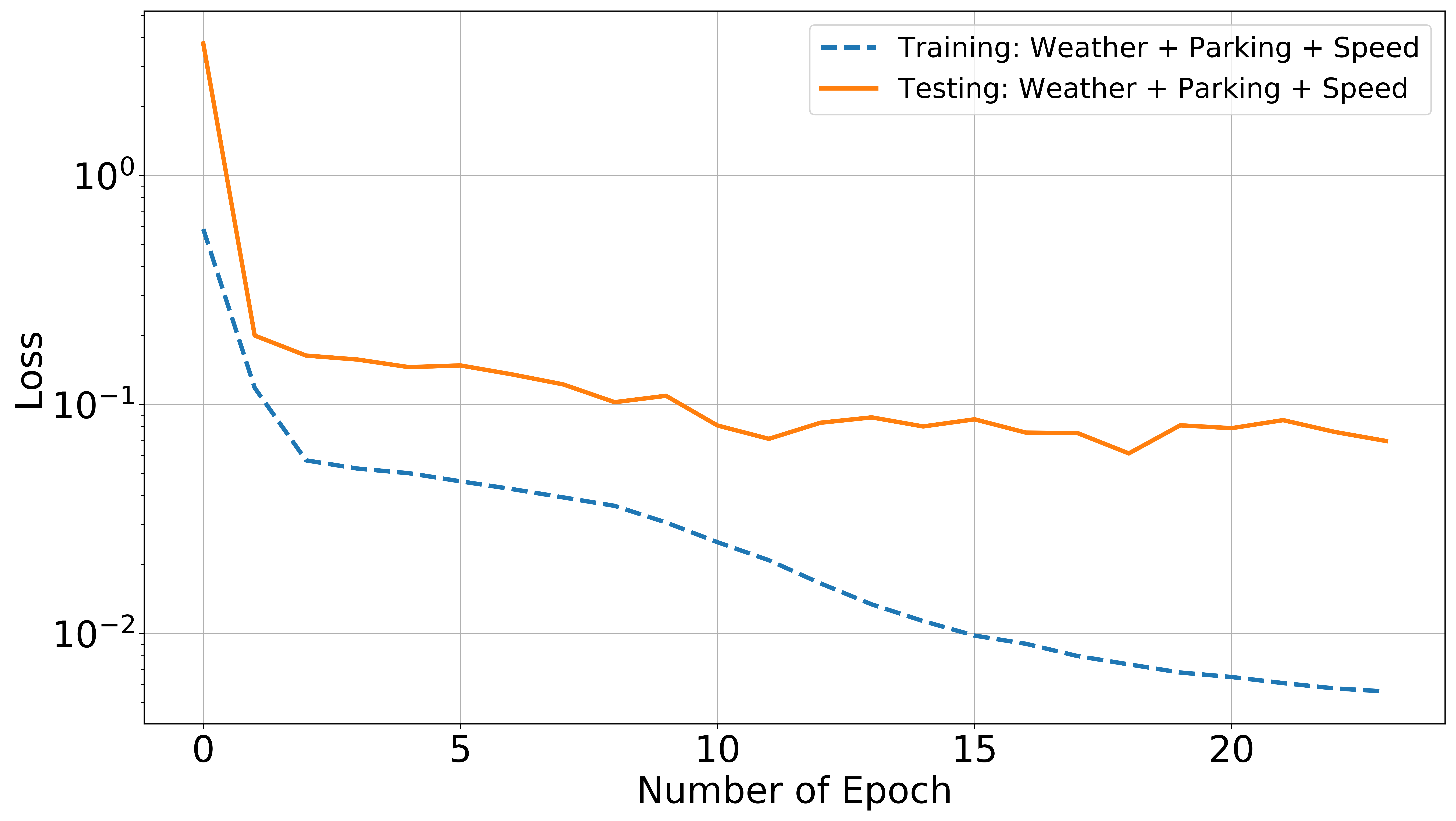}
	\caption{Model training  results}
	\label{fig:fitting}
\end{figure}

The training progress of the final model is presented in Fig. \ref{fig:fitting}. Its y-axis is the loss in log scale. Note that the loss is measured in Mean Square Error of the block-wise occupancies after preprocessing, i.e. occupancies values in the range of [-1, 1]. The proposed model converges quickly in the first 15 epochs. After that, the training loss continues to drop slowly while the testing loss starts to increase a bit as a sign of over-fitting. Finally, the training progress is terminated at the 23th epoch triggered by early stopping.


In the original scale of parking occupancy, i.e. the number of parked vehicles per block, the training and testing MAP of the model are 0.79 and 1.39, respectively. The average parking capacity across all 39 blocks is $14.25$, and the testing MAPE of the proposed model is 10.6\%. Due to the random nature of parking demand in urban areas, it is usually challenging to predict parking for a relatively long time window in advance, such as 30 min. As a result, most of the studies in the literature focused on small prediction windows, such as 5 min, \citet{vlahogianni2016real} evaluated their model 30 min in advance, but its for the regional level instead of the block level. For most parking applications on the current market, such as Google Maps and SFMTA\footnote{https://www.sfmta.com/demand-responsive-parking}, only the estimated parking occupancies or trends of historical average are offered instead of 30-min ahead prediction. 

One critical advantage of 30-min-ahead parking prediction over 5-min-ahead or real-time estimation is that the former would allow better trip planning for travelers, so that they can make proper decisions on their travel mode, whether they can drive and parking, or they need alternative modes to avoid cruising for parking. Parkers may not necessarily need to know the accurate rate of occupancy, but the probability of finding a vacant spot at the designated block (or namely the expected time spent on finding a parking spot) is useful and highly sensitive to the occupancy rate (the cruising time is the reciprocal of the number of vacant spots) \citep[e.g.,][]{qian2014optimal}. 30-min-ahead prediction at the block level could tell the travelers where exactly they should go for a vacant spot before they depart. Thus, marginally reducing the prediction error of occupancy rate, especially when it is close to being full (i.e., rate is 1), would be very effective for parking management. In general, a 10.6\% MAPE resulted from our model is capable of effectively identifying over-crowded parking blocks from those where traveler are likely to find a vacant spot by the time of arrival. Comparing to LASSO, the 25.9\% drop of MAPE from 14.3\% to 10.6\% is considered as a reasonable improvement in parking prediction.

\begin{figure}[H]
	\centering
	\includegraphics[width=0.9\textwidth]{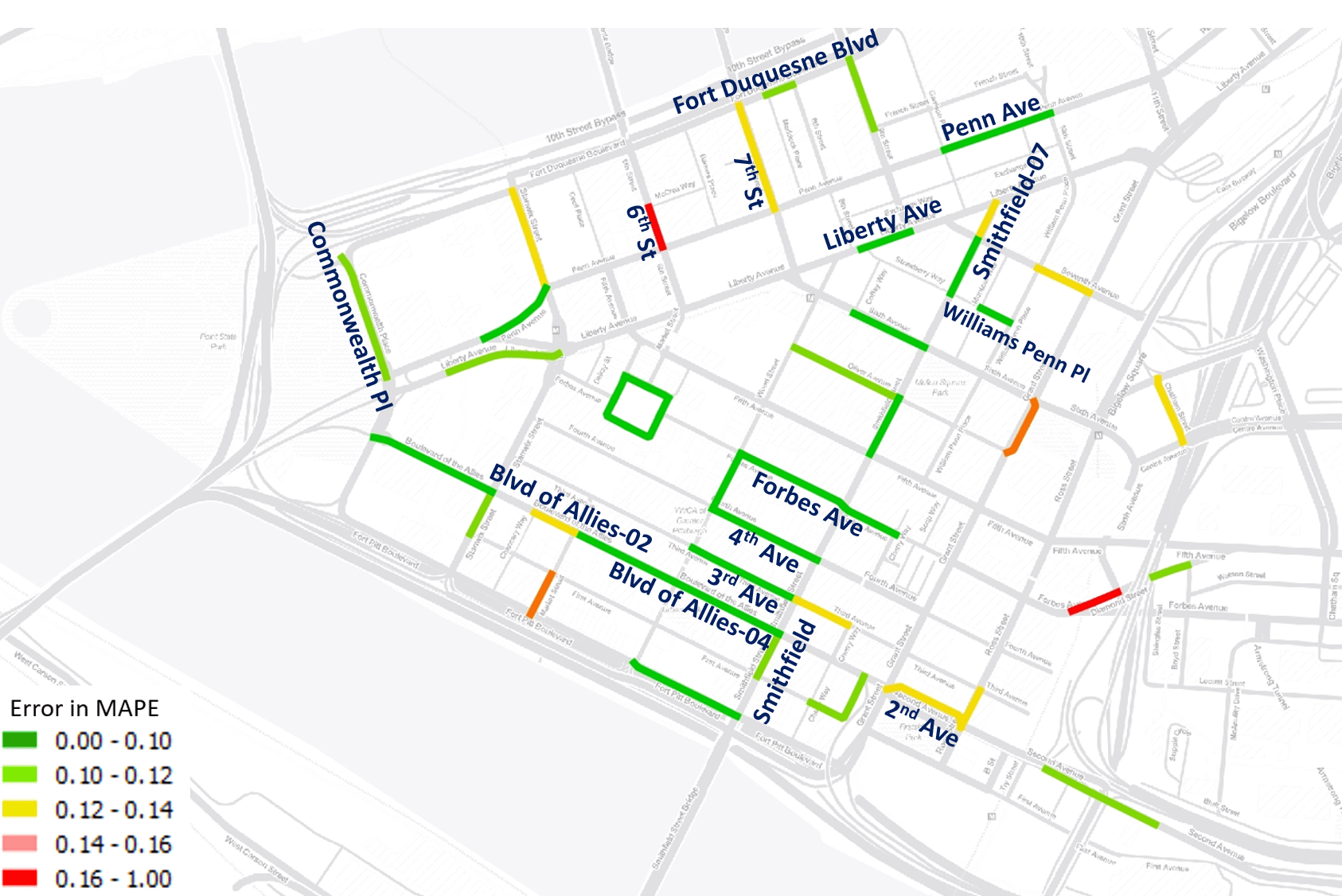}
	\caption{Heatmap of block-wise model performance (prediction errors are measured in MAPE)}
	\label{gcnn_heat}
\end{figure}

To evaluate and compare the performance of the proposed model in different locations, we visualize the block-level prediction error by a heatmap, as shown in Fig \ref{gcnn_heat}, especially, names of the parking blocks mentioned in this paper are attached on the Figure.  In general, lower prediction errors are received on blocks with larger parking capacities, such as 3rd Ave (28 spaces), 4th Ave (32 spaces); Liberty Ave (23 spaces) and Blvd of the Allies East side (23 spaces), with all MAPEs under 10\%. They are marked as dark green on the map. It is no surprise as higher parking capacities usually result in lower variances in occupancy rates. In addition, it is also found that the model performs better on business districts, such as Blvd of the Allies and Liberty Ave, while blocks with more recreational activities, such as 6th St and Commonwealth Place, are more difficult to predict. As previously discussed in Section \ref{data_ana}, parking demand in business districts is likely to be attributed to recurrent business activities which has strong daily and day-of-week patterns, comparing to recreational parking demand that is less likely to be recurrent from day to day. Demand patterns and characteristics in business districts are more resilient to impacts from unusual scenarios such as hazardous weather and special events. As a result, prediction performance for business districts are generally better than others.

\subsubsection{Comparison with baseline models}
In order to evaluate the effectiveness of the proposed method, we select several typical prediction models as baselines, including multi-layered LSTM, LASSO (least absolute shrinkage and selection operator), Historical Average and Latest Observation. To conduct a fair comparison, all baseline models are trained and tested on the same dataset as the proposed model. For the LSTM models, hyper-parameters including the dimension of each layer and the dropout rate are tuned by a grid search to ensure the best model configuration is adopted. Prediction performance is measured by MAE and MAPE of predicting block-level parking occupancies 30 minutes in advance.

For the 2-layered LSTM, the best result is reached on a configuration of using 1024 and 256 as the dimensions for the first and second layers, along with a dropout rate of 0.25; For the model of 3-layer LSTM, the best result is reached on a 2048-512-128 network configuration with a dropout rate of 0.25. As for LASSO, the objective function of optimization is shown in Eq. \ref{lassoequ},
\begin{equation}
\min_{\beta \in \mathbb{R}}\{\frac{1}{N}||y-X\beta||^2_2 + \lambda||\beta||_1\}
\label{lassoequ}
\end{equation}
Where $X, y, \beta$ being input features, occupancy ground truth and model parameters, respectively. In LASSO, all spatial information are dropped as the features being flattened into a one-dimensional vector. the optimization object of LASSO [not clear: fix]. In the Historical Average method, historical occupancy observations of the same time and week-of-day are averaged as the predicted value for individual blocks. In the Latest Observation method, the current occupancy is used as the predicted value for 30 minutes later.

Results of all models are shown in Table \ref{comparison}. The proposed network outperforms all baseline models with a significant margin. The closest one to our method is LASSO. Interestingly, the two multi-layer LSTM models are out-performed by LASSO, the linear regression model with Norm 1 constraints, indicating that some complex deep neural network models are prone to fail when working with high-dimensional but small-sized dataset. The GCNN, on the other hands, applies spectral filtering on the graph representation of the road network to capture the core spatial correlations, while bounding the model complexity by parameter sharing. Thus is more suitable for scenarios with strong spatial correlations among features than other deep learning models like vanilla RNN.

\begin{table}[H]
	\centering
	\caption{Performance comparisons between models}
	\begin{tabular}{|c|c|c|}
		\hline
		{\bf Model}             & {\bf Test MAE} & {\bf Test MAPE} \\ \hline
		GCNN+LSTM (proposed model)     & 1.39  & 10.6\% \\ \hline
		2-layer LSTM         & 1.87   & 14.6\%  \\ \hline
		3-layer LSTM        & 2.07   & 15.7\%  \\ \hline
		LASSO               & 1.82  & 14.3\%   \\ \hline
		Historical Average  & 3.49   & 25.4\%  \\ \hline
		Latest Observation     & 2.66   & 19.9\%  \\ \hline
	\end{tabular}
	\label{comparison}
\end{table}

To provide a more straightforward illustration of the predicted occupancies on various locations, an example output of prediction models at a particular tisme interval is presented in table \ref{sample_output}. Five blocks with a wide spectrum of parking capacities are selected as representative examples. The parking capacity, true occupancy and the predicted values for 2014/12/04 13:00 EST are listed in the table. Among them, Smithfield-07 is the northmost segment of Smithfield, Blvd of Allies-02 is the second westmost segment of Blvd of Allies that allows on-street parking, between Market St and Wood St. Although none of the models are perfect in this case, the proposed GCNN+LSTM model outperforms other baseline models as this is generally the case. For this specific time interval, occupancies predicted by historical average fail to represent the true parking conditions, especially on Fort Duquesne and Smithfield-07, while models such as GCNN or LASSO are capable of forecasting relatively low parking occupancies on these two blocks that is close to the ground truth.

\begin{figure}[H]
	\centering
	\includegraphics[width=0.9\textwidth]{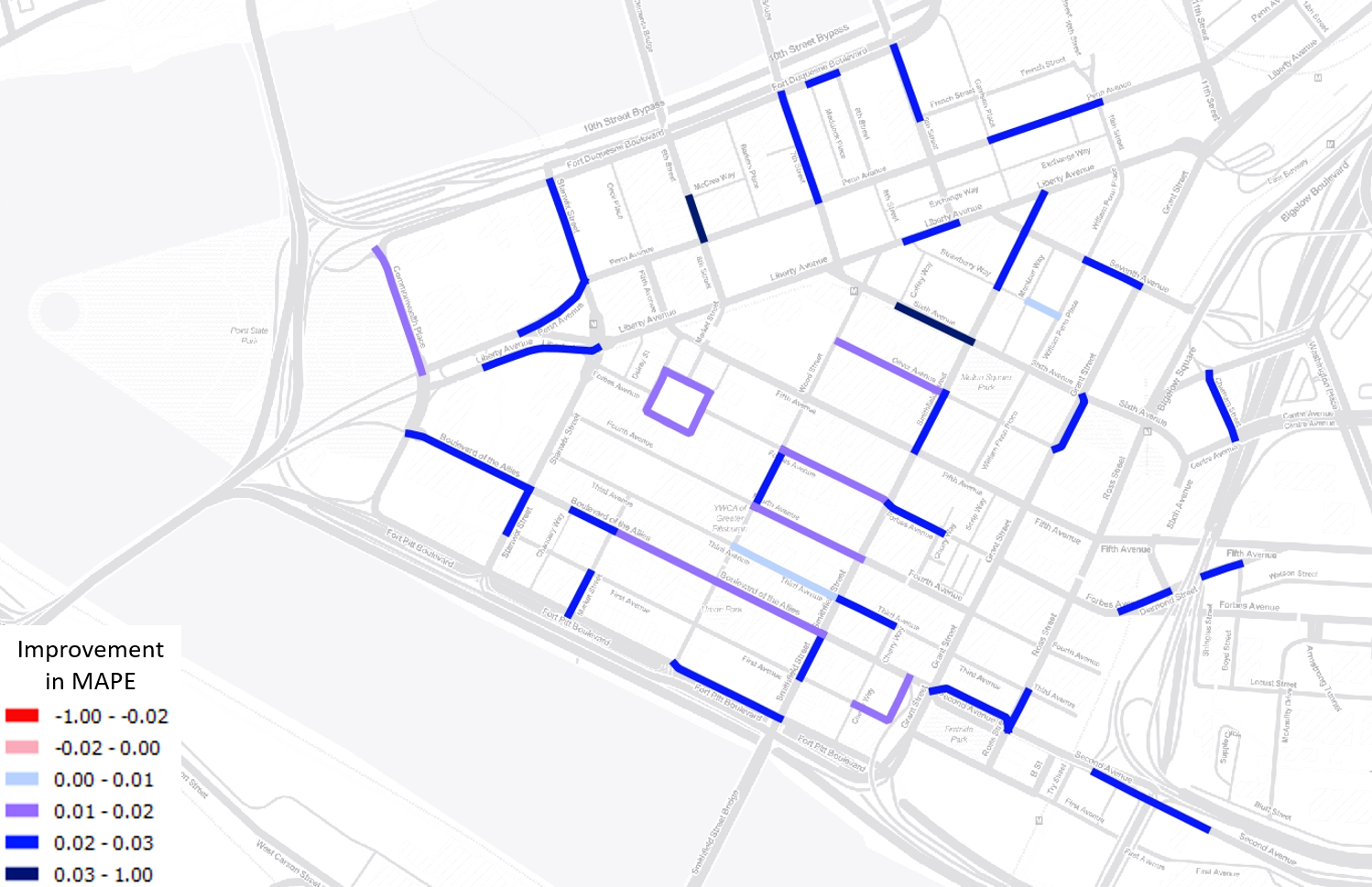}
	\caption{Heatmap of block-level comparison of performance between proposed model and LASSO, the value of each block is $E_{LASSO} - E_{GCNN}$, where $E_{LASSO}, E_{GCNN}$ are the prediction errors in MAPE of LASSO and the proposed model respectively}
	\label{LASSO_heat}
\end{figure}

\begin{table}[H]
	\centering
	\caption{Sample output of predicted occupancies on five blocks at 2014/12/04 13:00 EST}
	\begin{tabular}{|c|c|c|c|c|c|}
		\hline
		& Fort Duquesne & Smithfield-07 & 7th St & Blvd of Allies-02 & 2nd Ave \\ \hline
		Parking capacity   & 25            & 16            & 8      & 27                & 29      \\ \hline
		True occupancy     & 6             & 4             & 6      & 22                & 23      \\ \hline
		GCNN+LSTM          & 4.75          & 6.12          & 5.74   & 22.95             & 26.25   \\ \hline
		2-layer LSTM       & 3.96          & 4.98          & 7.02   & 17.56             & 22.28   \\ \hline
		3-layer LSTM       & 4.51          & 5.02          & 7.31   & 19.41             & 20.99   \\ \hline
		LASSO              & 3.90          & 6.17          & 7.43   & 18.87             & 25.14   \\ \hline
		Historical Average & 18.06         & 11.11         & 4.26   & 17.54             & 16.93   \\ \hline
		Last Observation   & 4             & 3             & 2      & 10                & 17      \\ \hline
	\end{tabular}
	\label{sample_output}
\end{table}
Furthermore, we also compare the prediction efficiency between the proposed model and LASSO for each of the 39 blocks in this study. The heatmap of block-level performance improvement are shown in Fig \ref{LASSO_heat}, where the value of each block is $E_{LASSO} - E_{GCNN}$, where $E_{LASSO}, E_{GCNN}$ are the prediction errors in MAPE of LASSO and the proposed model respectively. The proposed model outperforms LASSO in all 39 parking blocks over the Pittsburgh downtown area. We found that the actual performance discrepancies vary by location, but the proposed model generally improves results more over LASSO in several blocks with lower parking capacities. This is consistent with findings before, blocks with high parking capacities are relatively easier to predict. Incorporating spatial correlations is generally more effective for blocks with higher variances of occupancies, as the percentage occupancies of these blocks will bear greater impacts when their neighboring blocks are full, and the subsequent parking demand ``spills back'' to them.

On the other hand, the standard deviation of the block-level MAPE of the proposed model is 0.0252, comparing to 0.0381 using LASSO. In other words, LASSO’s standard deviation is 47\% higher than the proposed method. In conjunction with the observations from Fig 10, we conclude that, in general, the proposed deep learning based model is  more robust and adaptive to different types of parking locations.

\subsubsection{The role of each data source}
\begin{figure}[H]
	\centering
	\begin{subfigure}[b]{0.9\textwidth}
		\includegraphics[width=\textwidth]{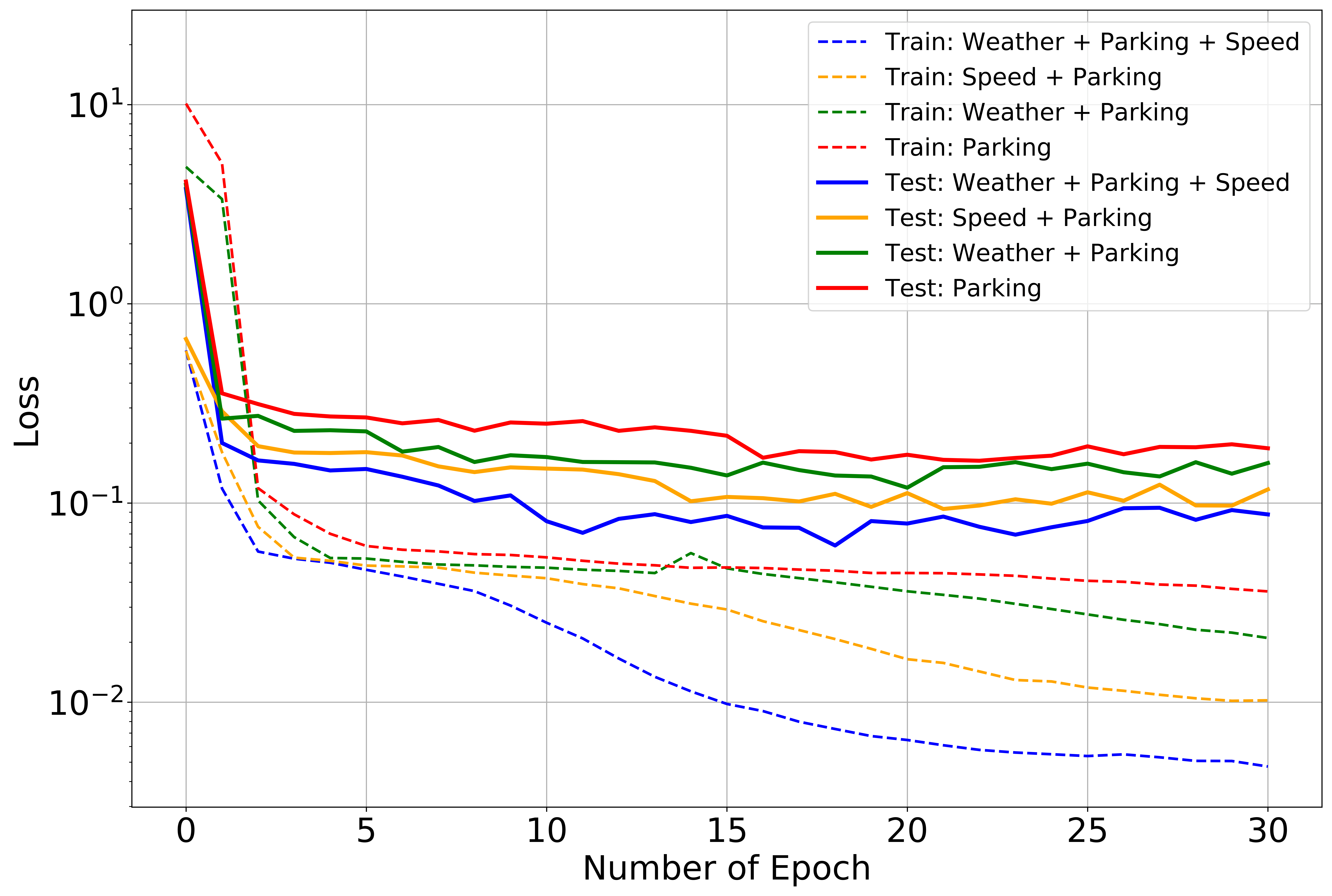}
	\end{subfigure}
	\caption{Convergence profile for different combinations of data sources}
	\label{fig:com}
\end{figure}

To evaluate the contributions of each data source in parking occupancy prediction, we conduct experiments on different combinations of data sources. Thanks to the flexibility of the model framework, this part can be easily done with no twist on the framework. As the parking transactions are the major source of parking occupancy information, it is kept in all experiments. The other two sources, traffic speed and weather, are included or excluded from the model to create different feature space. The training and testing results are presented in Fig. \ref{fig:com}.

The original model with all data sources outperforms the three other models that use only partial data. The model with parking and traffic speed reaches the second best result. The speed profile of the network reflects the real-time demand and congestion of both parking and traffic flow. This confirms our assumption that the correlation between traffic congestion and parking occupancy is significant. Also, the spatio-temporal traffic congestion pattern can be a good indicator for non-recurrent travel and parking demand surge, which is not directly reflected through parking transaction data (even with a time lag). Weather features also contribute to the effectiveness of the parking occupancy prediction, despite not as much as the traffic speed data. Weather may be partially explained already through the traffic speed as they are likely correlated. However, results show weather data can add additional value to the parking occupancy prediction compared with traffic speed.

\begin{figure}[H]
	\centering
	\includegraphics[width=0.8\textwidth]{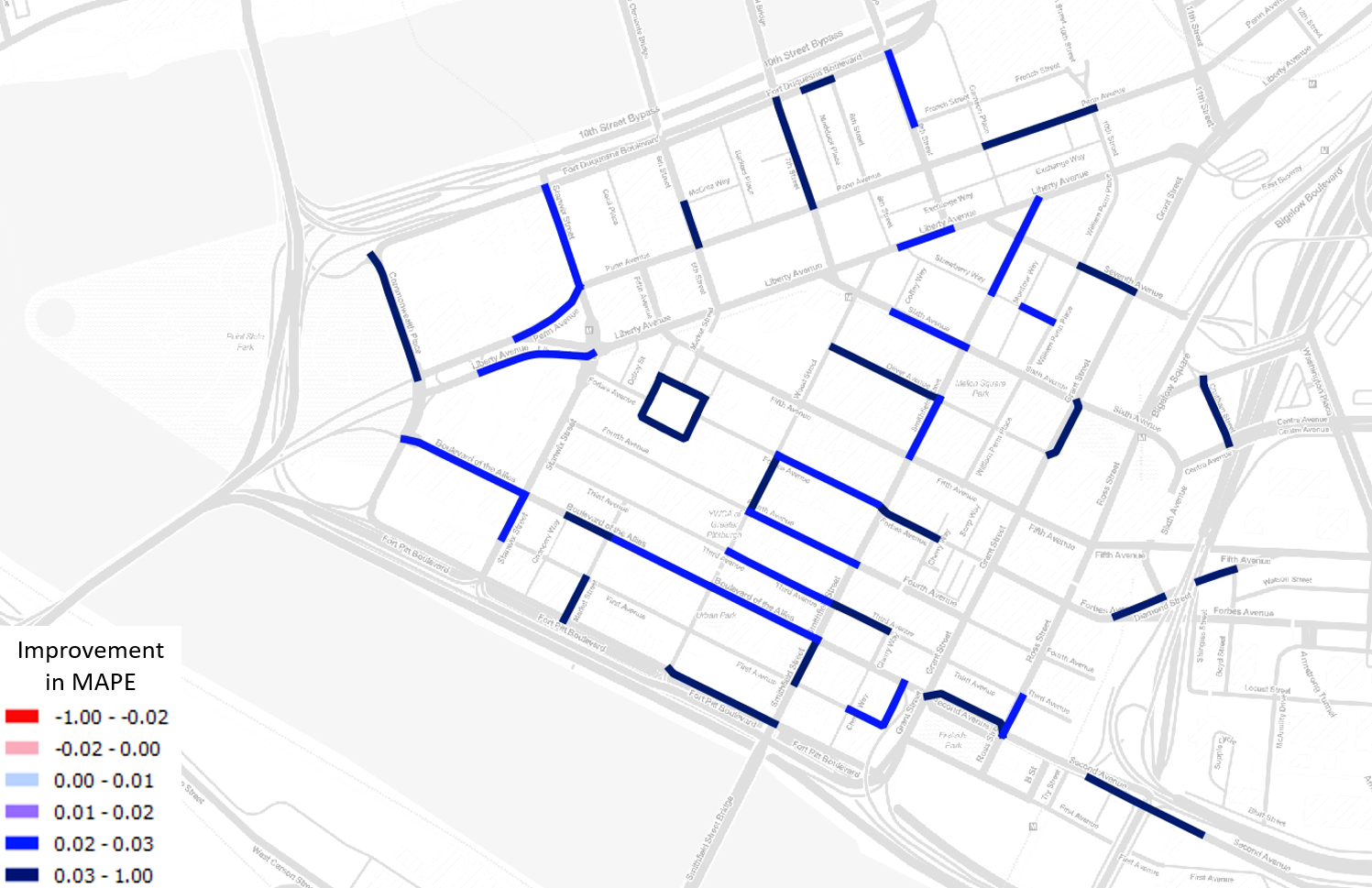}
	\caption{Heatmap of block-level performance improvement by incorporating traffic speed information (errors are measured in MAPE)}
	\label{gcnn_no_traffic}
\end{figure}

\begin{figure}[H]
	\centering
	\includegraphics[width=0.8\textwidth]{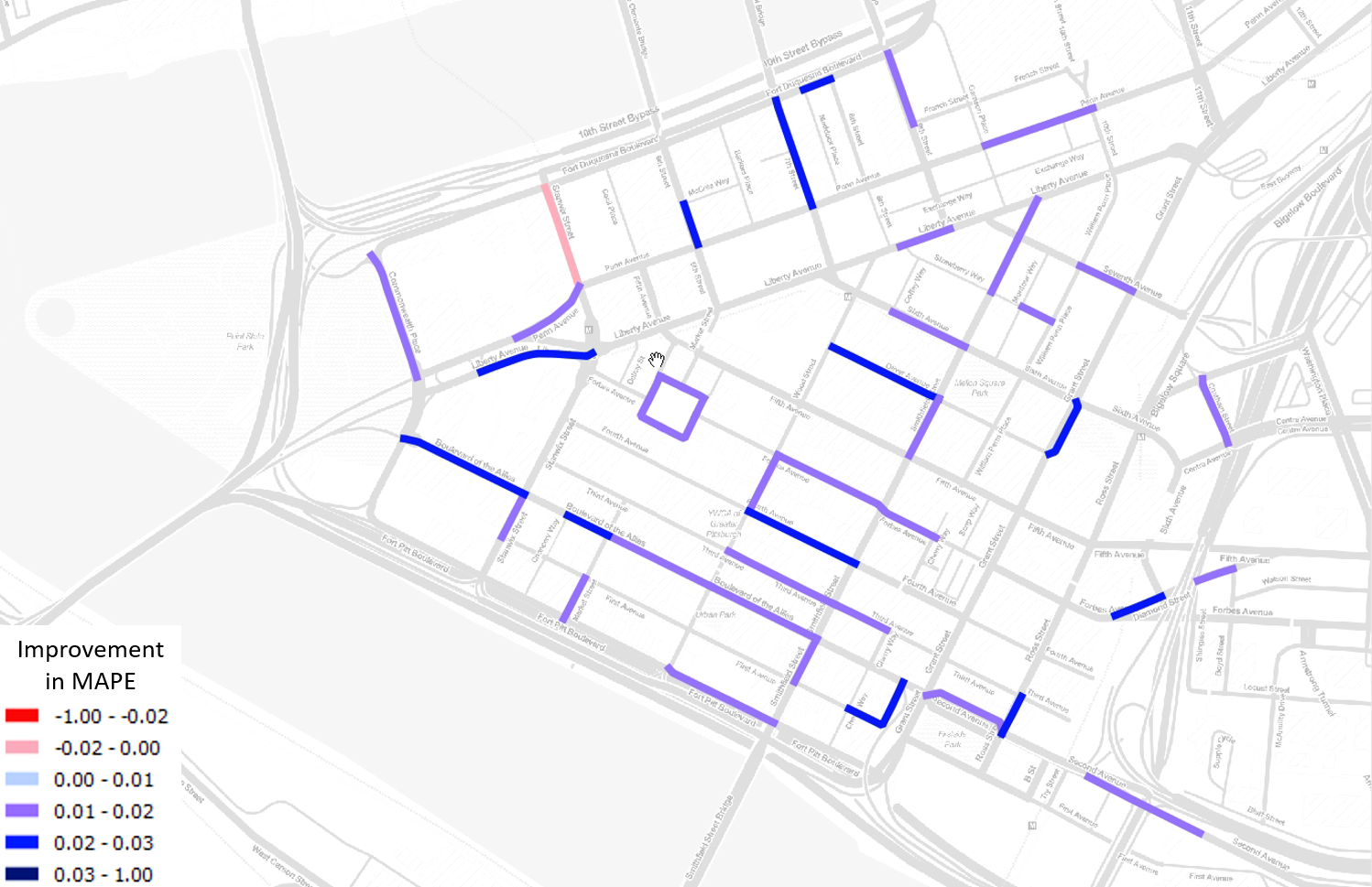}
	\caption{Heatmap of block-level performance improvement by incorporating weather conditions (errors are measured in MAPE)}
	\label{gcnn_no_weather}
\end{figure}

We also visualize the block-level performance improvement from the incorporation of speed and weather dataset, as shown in Fig \ref{gcnn_no_traffic} and Fig \ref{gcnn_no_weather}, respectively. Note that the two figures in this subsection and the figure implying the improvement over LASSO in Fig \ref{LASSO_heat} share the same legends. We can see that the performance gains brought by traffic speed information are substantial across all blocks, ranging from 1.5\% to 4.4\% in MAPE. In Section \ref{data_ana} we have discussed that the relationship between speed and block-level parking occupancies are complex and non-linear, with an example of the correlations on Penn Ave East and Second Ave in Fig \ref{fig:spd2blk}. As a comparison, the improvements led by speed data are more than 3\% for both of the two locations. Thus, by properly incorporating the speed data, the performance of occupancy predictions can be improved by a significant margin. In addition, the proposed method, combining graph convolution with LSTM, is an efficient way to model the complex relationship between speed and occupancies.

The block-level performance improvement from weather information is visualized by a heatmap in Fig \ref{gcnn_no_weather}. The prediction errors decline in 38 out of the 39 blocks after incorporating weather features, up to 3.0\% in MAPE. In corresponding to the results in Fig \ref{fig:weather2blk}, the improvement on Liberty Ave is 1.1\%, and 1.8\% on Commonwealth Pl. Generally, weather features are more useful for recreational locations than business areas, since the parking demand in the latter is less sensitive to hazardous weather conditions.


\section{Conclusions} \label{cond}
This paper proposes a deep learning model for block-level parking occupancy prediction in large-scale networks. The model incorporates Graph CNN, LSTM and FNN, and has the flexibility to take multiple spatial-temporal structured data sources as input. Performance of the model is evaluated on a case study in the Pittsburgh downtown area, in which parking meter transactions, traffic speed, and weather data along with road network data are used. The model outperforms other baseline methods including multi-layer LSTM and LASSO by a significant margin. In generally, the prediction model works better for business areas than for recreational locations. We found that incorporating traffic speed and weather information that potentially influence parking behavior can significantly improve the performance of parking occupancy prediction. Weather data is particularly useful for improving predicting accuracy in recreational areas.

A reliable prediction of parking occupancy far ahead is critical to proactive real-time parking management,
\begin{enumerate}
	\item Effective parking guidance. Only if we assess the parking conditions ahead of time, we can then provide proper guidance to the incoming drivers. The key advantage of predicting in 30 min advance rather than just 10 or 5 min in some of the past literature is that it allows better trip planning for travelers, so that they can make proper decisions on their travel mode, whether they can drive and parking, or they need alternative modes to avoid cruising for parking. Also, practically, knowing where they can find parking spaces 30 min ahead can greatly reduce the need of using the parking guidance app on mobile phones while driving.
	\item Dynamic parking rates and parking enforcement. If we have reliable sources of predicted parking occupancy in the near future, we can apply dynamic parking pricing to control the parking demand for each parking blocks, as well as the total travel demand in the road network. Predicting parking occupancies far ahead is more critical than estimating parking occupancies in real time. In the latter case, parkers may arrive and find the rates too high to park, which can result in unnecessary cruising for parking and additional roadway congestion.”
\end{enumerate}

The prediction method can be further improved in the following ways,
\begin{enumerate}
	\item Redefine the spatial unit of the occupancy prediction and try different methods for incorporating the road network topology into the prediction. We currently perform the prediction for each block, where each block is essentially a single or several road segments. In the near future, we plan to apply similar methodology to occupancy prediction at multiple scales, from the level of each parking meter to the level of aggregated zones.
	\item Infer causal relationships of parking occupancy among different parking blocks. For instance, how is parking occupancy propagated through the network.
	\item Conduct feature selection and network pruning, which could further mitigate the issue of over-fitting and improve the model performance.
	\item Collect and evaluate the effectiveness of additional traffic-related data, including traffic counts, road closure, incidents, and events.
	\item Run our algorithms on street blocks that have set up dynamic parking rates (this was implemented in San Francisco, Seattle, and some parts of Pittsburgh). We will incorporate the impact of pricing on parking demand/occupancy in the prediction model by collecting sufficient data under the dynamic pricing scheme and re-training the prediction model that adapts to the pricing effect.
\end{enumerate}

\section*{Acknowledgements}
This research is funded in part by National Science Foundation Award CNS-1544826 and Carnegie Mellon University's Mobility21, a National University Transportation Center for Mobility sponsored by the US Department of Transportation. The contents of this report reflect the views of the authors, who are responsible for the facts and the accuracy of the information presented herein. The U.S. Government assumes no liability for the contents or use thereof.

\bibliographystyle{elsarticle-harv}
\bibliography{parking}
\end{document}